%% file: pami-main.tex
\documentclass[lettersize,journal]{IEEEtran}
\usepackage{amsmath,amsfonts}
\usepackage{algorithmic}
\usepackage{algorithm}
\usepackage{array}
\usepackage[caption=false,font=normalsize,labelfont=sf,textfont=sf]{subfig}
\usepackage{textcomp}
\usepackage{stfloats}
\usepackage{url}
\usepackage{verbatim}
\usepackage{graphicx}
\usepackage{cite}
\usepackage{booktabs}
\usepackage{color}
\usepackage{amsmath}
\usepackage{xcolor}
\usepackage{tabularx}
\usepackage[breaklinks=true, hidelinks]{hyperref}

\begin{document}

\title{Wonder3D++: Cross-domain Diffusion for High-fidelity 3D Generation from a Single Image}

\author{Yuxiao Yang$^{\ast}$, Xiao-Xiao Long$^{\ast}$, Zhiyang Dou, Cheng Lin, Yuan Liu, Qingsong Yan, \\ Yuexin Ma, 
Haoqian Wang,  Zhiqiang Wu$^{\dagger}$, Wei Yin$^{\dagger}$
\thanks{Yuxiao Yang and Xiao-Xiao Long contribute equally to this work. Yuxiao Yang is with Tsinghua University and also with Nanjing University. Xiaoxiao Long is with Nanjing University.} 
\thanks{Zhiqiang Wu and Wei Yin are the corresponding authors. Zhiqiang Wu is with Wright State University. Wei Yin is with Horizon Robotics. Email: zhiqiang.wu@wright.edu, yvanwy@outlook.com}
\thanks{Zhiyang Dou is with the University of Hong Kong.}
\thanks{Cheng Lin is with the Macau University of Science and Technology.}
\thanks{Yuan Liu is with the Hong Kong University of Science and Technology.}
\thanks{Qingsong Yan is with Wuhan University.}
\thanks{Yuexin Ma is with ShanghaiTech University.}
\thanks{Haoqian Wang is with Tsinghua University.}
}



\maketitle
\input{figures/teaser}
\input{sections/0_abs}


\input{sections/1_intro}

\input{sections/2_related_works}

\input{sections/3_method}

\input{sections/5_exp}

\input{sections/6_conclusion}

\input{sections/acknowledgment}

{
\bibliographystyle{plain}
\bibliography{pami-main}
}

\vfill

\input{sections/bio}

\clearpage

\appendices
\input{sections/7_supp}

\end{document}

%% file: figures/teaser.tex
\begin{figure*}[tp!]
\centering
\includegraphics[width=\linewidth]{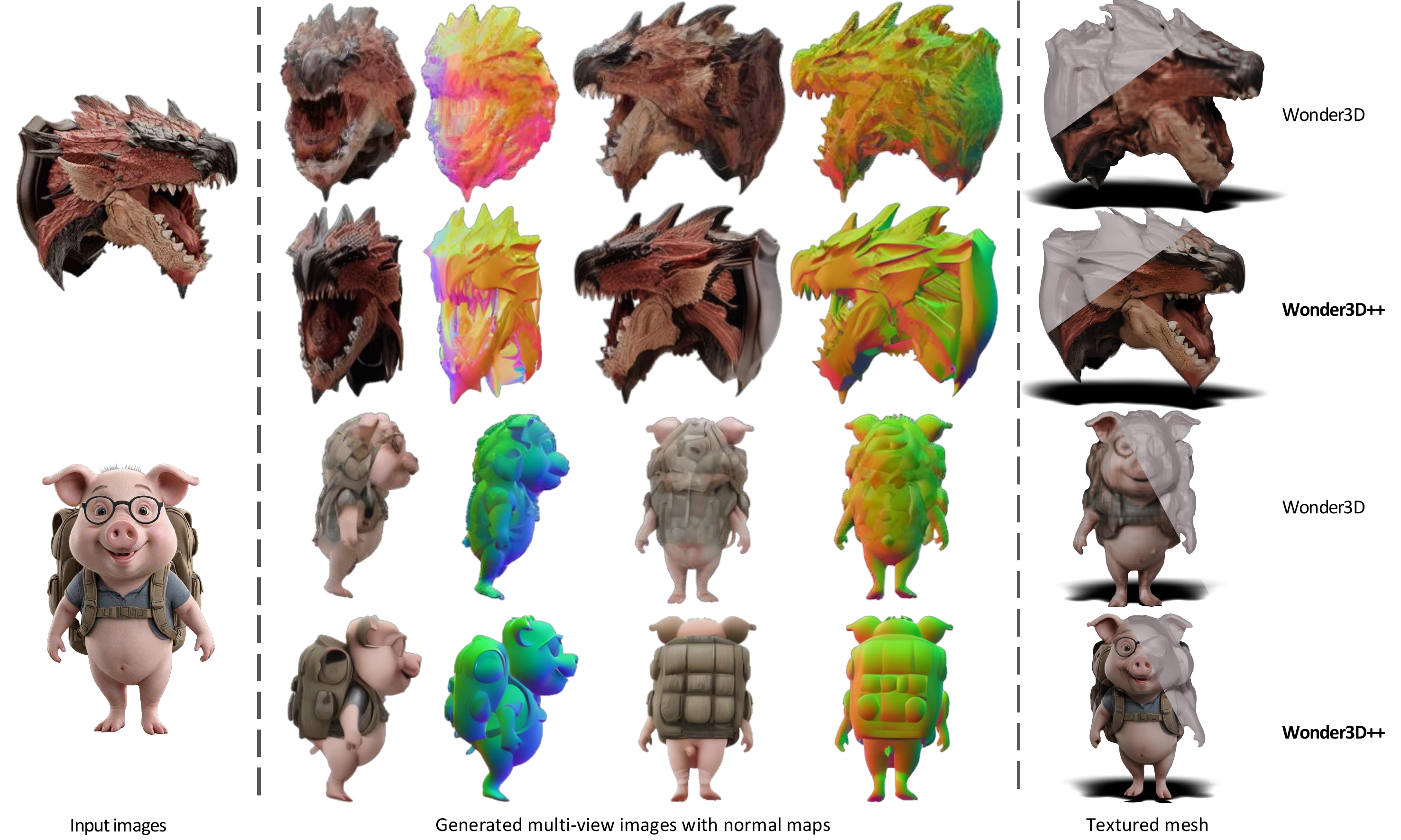}
\caption{\textit{Wonder3D++} reconstructs highly-detailed textured meshes from a single-view image in only 3 minutes. \textit{Wonder3D++} first generates consistent multi-view normal maps with corresponding color images via a cross-domain diffusion model and then leverages a cascaded 3D mesh extraction method to achieve fast and high-quality reconstruction. 
}
\label{fig:teaser}	
\end{figure*}

%% file: sections/0_abs.tex
\begin{abstract}

In this work, we introduce \textbf{Wonder3D++}, a novel method for efficiently generating high-fidelity textured meshes from single-view images.
Recent methods based on Score Distillation Sampling (SDS) have shown the potential to recover 3D geometry from 2D diffusion priors, but they typically suffer from time-consuming per-shape optimization and inconsistent geometry. 
In contrast, certain works directly produce 3D information via fast network inferences, but their results are often of low quality and lack geometric details.
To holistically improve the quality, consistency, and efficiency of single-view reconstruction tasks, we propose a cross-domain diffusion model that generates multi-view normal maps and the corresponding color images. To ensure the consistency of generation, we employ a multi-view cross-domain attention mechanism that facilitates information exchange across views and modalities. Lastly, we introduce a cascaded 3D mesh extraction algorithm that drives high-quality surfaces from the multi-view 2D representations in only about $3$ minute in a coarse-to-fine manner.
Our extensive evaluations demonstrate that our method achieves high-quality reconstruction results, robust generalization, and good efficiency compared to prior works. Code available at \url{https://github.com/xxlong0/Wonder3D/tree/Wonder3D_Plus}.
 
\end{abstract}

%% file: sections/1_intro.tex
\section{Introduction}
Reconstructing 3D geometry from a single image~\cite{jun2023shap, melas2023realfusion, dou2023tore, liu2023zero,Long_2021_ICCV, nichol2022point,long2024adaptive} stands as a fundamental task in computer graphics and 3D computer vision,
benefiting a wide range of versatile applications such as novel view synthesis~\cite{liu2022neural,jiang2023gaussianshader,cheng2024gaussianpro}, 3D content creation~\cite{liu2023syncdreamer,qian2023magic123}, and robotics grasping~\cite{zheng2024gaussiangrasper,kleeberger2020survey}.
However, this task is notably challenging since it is ill-posed and demands the ability to discern the 3D geometry of both visible and invisible parts. This ability requires extensive knowledge of the 3D world.

Recently, the field of 3D generation has experienced rapid and flourishing development with the introduction of diffusion models. A growing body of research~\cite{poole2022dreamfusion,wang2023score,wang2023prolificdreamer,lin2023magic3d,chen2023fantasia3d}, such as DreamField~\cite{jain2022zero}, DreamFusion~\cite{poole2022dreamfusion}, and Magic3D~\cite{lin2023magic3d}, resort to distilling prior knowledge of 2D image diffusion models or vision language models to create 3D models from text or images via Score Distillation Sampling (SDS)~\cite{poole2022dreamfusion}. 
Despite achieving some degree of generalization performance, these methods still encounter challenges like multi-view inconsistencies and low-quality results.

To enhance the multi-view consistency, several methods emerged that directly generate multi-view 2D images, with representative works including SyncDreamer~\cite{liu2023syncdreamer} and MVDream~\cite{shi2023mvdream}. 
By improving information exchange between views using attention mechanisms, existing methods generate multi-view images and then optimize 3D representations—such as Neus~\cite{wang2021neus} and NeRF~\cite{mildenhall2020nerf}—to reconstruct textured meshes. However, current multi-view generative methods face significant challenges. First, relying solely on color images often compromises the fidelity of the generated shapes, making it difficult to recover geometric details without incurring high computational costs. Additionally, most multi-view diffusion models are constrained to predefined camera types, which can introduce distortions when processing images captured in uncontrolled environments, further amplifying multi-view inconsistencies.

Recently, various techniques have been introduced that aim to generate 3D geometries via network inference, including point clouds~\cite{nichol2022point, zeng2022lion, luo2021diffusion, zhou20213d}, meshes~\cite{liu2023meshdiffusion,gao2022get3d}, neural fields~\cite{wang2023rodin,cheng2023sdfusion,karnewar2023holofusion, kim2023neuralfield, gu2023learning, anciukevivcius2023renderdiffusion, muller2023diffrf, ntavelis2023autodecoding, jun2023shap, zhang20233dshape2vecset, gupta20233dgen, erkocc2023hyperdiffusion, chen2023single, hong2023lrm}. Most of them attempt to train 3D generative models from scratch on datasets consisting of various 3D assets. 
To further enhance generalizability, recent approaches have focused on generating sparse view images using a multi-view diffusion model as a supplementary cue, followed by a feed-forward sparse-view reconstruction model to create 3D objects~\cite{tang2024lgm, wang2024crm, xu2024grm, xu2024instantmesh}. However, despite some of these methods requiring over a hundred GPUs for training, they still face significant challenges in achieving high-quality geometry and textures. Additionally, since these methods rely on neural networks for direct forward propagation to generate 3D models, the resulting objects often lack precise alignment with the input images.
\IEEEpubidadjcol

To address the challenges of fidelity, consistency, and generalizability in previous works, we propose a novel approach for single-view 3D reconstruction by generating multi-view consistent normal maps and their corresponding color images using a cross-domain diffusion model. The core idea is to extend diffusion frameworks to jointly model the distribution of two domains: normals and colors. Normal maps capture surface undulations and variations, encoding detailed geometric information that allows for high-fidelity 3D geometry extraction from 2D representations.
By leveraging 2D representations, \textit{Wonder3D++} is built on the pre-trained Stable Diffusion model~\cite{radford2021learning}, utilizing strong priors that enable robust zero-shot generalization.

The following technical designs of \textit{Wonder3D++} make it a robust and efficient tool for creating 3D shapes from single images:
1) \textbf{Cross-domain switcher}: The domain switcher enables the diffusion model to generate either normal maps or color images with minimal modifications to the original model.
2) \textbf{Cross-domain attention}: We incorporate cross-domain attention mechanisms to facilitate information exchange between the two domains, improving consistency and quality. This enhances the model’s ability to recover high-fidelity geometry.
3) \textbf{Camera type switcher}: To handle input images from various sources—such as imaginative images generated by text-to-image models or everyday objects captured by smartphones—we introduce a camera type switcher. This allows the model to generate multi-view images using either perspective or orthogonal projection, enhancing its robustness.
4) \textbf{Cascaded 3D mesh extraction}: To address resolution limitations and improve efficiency, we propose a Cascaded 3D mesh extraction algorithm. This consists of a) geometric initialization, b) inconsistency-aware coarse reconstruction, and c) iterative mesh refinement via our cross-domain multi-view enhancement module. This process generates high-quality textured meshes while mitigating issues of multi-view inconsistency and limited resolution (see Figure~\ref{fig:teaser}).

We conduct extensive experiments on the Google Scanned Object dataset~\cite{downs2022google} and various 2D images with different styles.
The experiments validate that \textit{Wonder3D++} achieves a leading level of geometric details with high efficiency and strong generalization among current zero-shot single-view reconstruction methods.

\input{figures/pipeline}

\noindent
\textbf{Differences with conference version.}
\textit{Wonder3D++} introduces several key improvements over \textit{Wonder3D}~\cite{long2024wonder3d}, significantly enhancing robustness and quality: 1) the addition of a \textbf{camera type switcher}, allowing the model to handle input images captured with various camera settings; 2) the implementation of \textbf{multi-domain pre-training}, ensuring a seamless transition from stable diffusion pre-training to the desired joint distribution for 3D data; and 3) replacing implicit neural SDF fields with explicit mesh representations for 3D textured mesh reconstruction, introducing a \textbf{cascaded 3D mesh extraction} algorithm that greatly improves both efficiency and effectiveness.

%% file: figures/pipeline.tex
\begin{figure*}[tp!]
\centering
\includegraphics[width=\linewidth]{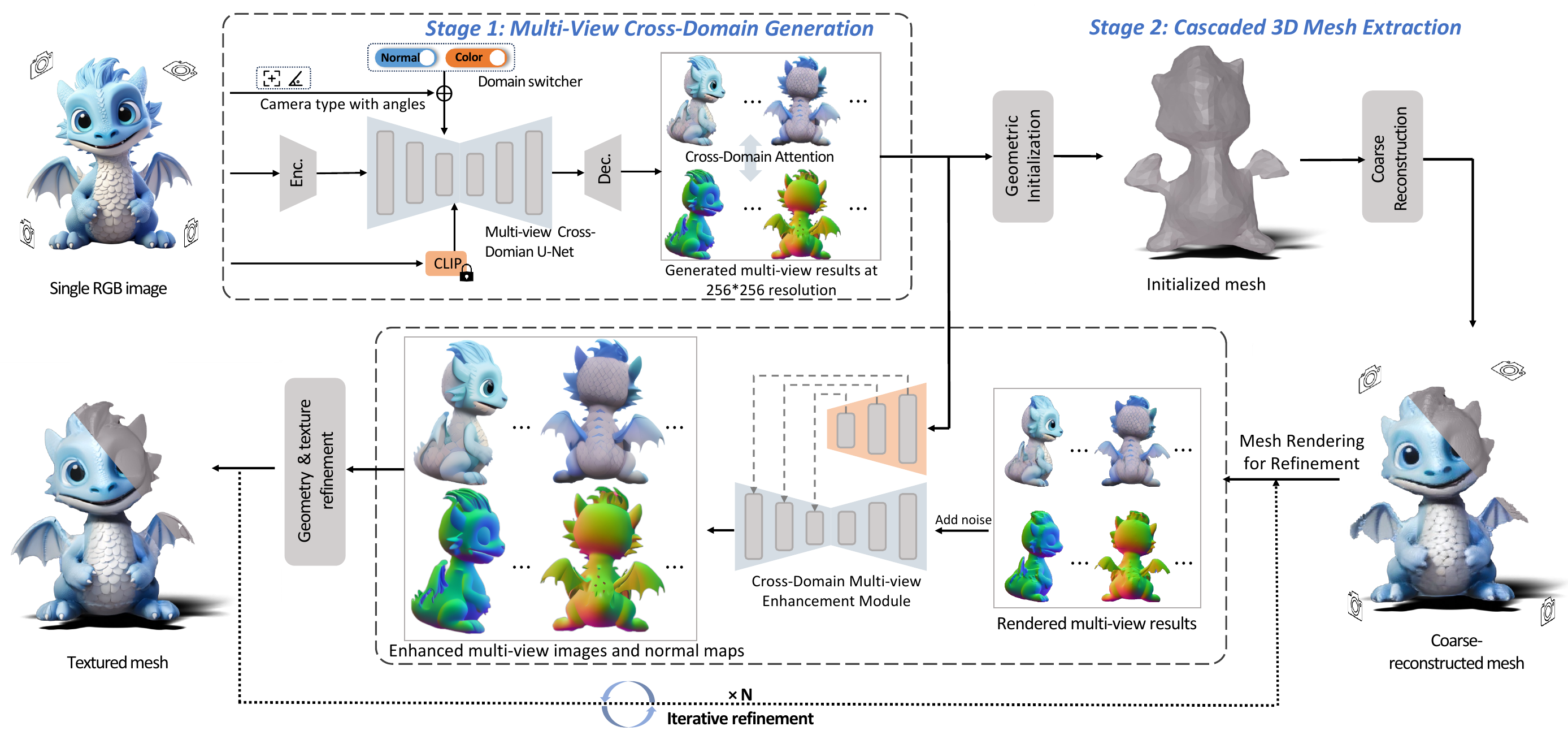}
\caption{Overview of \textit{Wonder3D++}. Given a single image, \textit{Wonder3D++} takes the input image feature produced by VAE encoder, the image embedding produced by CLIP model~\cite{radford2021learning}, the camera parameters of multiple views, a camera type switcher and a domain switcher as conditioning to generate consistent multi-view normal maps and color images. Subsequently, \textit{Wonder3D++} applies an innovative cascaded 3D mesh reconstruction algorithm, which utilizes a coarse-to-fine strategy consisting of geometric initialization, coarse mesh reconstruction, and iterative mesh refinement to produce a 3D mesh with high-quality geometry and high-fidelity textures. 
}
\label{fig:pipeline}	
\end{figure*}

%% file: sections/2_related_works.tex
\section{Related Works}
\subsection{2D Diffusion Models for 3D Generation}

Recently, benefiting from the availability of large-scale high-quality data, there have been significant advancements in foundational models, such as 2D diffusion models~\cite{ho2020denoising,rombach2022high,croitoru2023diffusion} and large vision-language models (e.g., the CLIP model~\cite{radford2021learning}). 
These models offer new opportunities for generating 3D assets by leveraging the strong priors provided by these pre-trained architectures.
In comparison, directly trained native 3D generators often struggle to ensure both generalizability and quality, due to the relative scarcity of large-scale data in the three-dimensional domain. 
Therefore, pioneering works such as DreamFusion~\cite{poole2022dreamfusion} and SJC~\cite{wang2023score} have proposed distilling a 2D text-to-image generation model to generate 3D shapes from textual descriptions, and subsequent works have adopted this per-shape optimization scheme, building upon the success of these initial approaches.
For the task of text-to-3D~\cite{chen2023fantasia3d,wang2023prolificdreamer,seo2023ditto,yu2023points,lin2023magic3d,seo2023let,tsalicoglou2023textmesh,zhu2023hifa,huang2023dreamtime,armandpour2023re,wu2023hd,chen2023it3d} or image-to-3D synthesis~\cite{tang2023make,melas2023realfusion,qian2023magic123,xu2022neurallift,raj2023dreambooth3d,shen2023anything}, these methods typically optimize a 3D representation (i.e., NeRF, mesh, or SDF), and then leverage differentiable rendering to deliver 2D images from various viewpoints. The images are then fed into the 2D diffusion models or CLIP model for calculating SDS~\cite{poole2022dreamfusion} losses, which guide the 3D shape optimization.
While this endows 3D generation methods with unprecedented generalization capabilities, most of these methods suffer from low efficiency and the “multi-face” problem (i.e., the Janus problem~\cite{poole2022dreamfusion}) due to the lack of explicit 3D supervision. Moreover, per-shape optimization, which typically takes tens of minutes, significantly limits the practicality of this methodology.

\subsection{Multi-view Diffusion Models}
To alleviate the "multi-face" problem, some efforts~\cite{watson2022novel,gu2023nerfdiff,deng2023nerdi,zhou2023sparsefusion,tseng2023consistent,chan2023generative,yu2023long,tewari2023diffusion,yoo2023dreamsparse,szymanowicz2023viewset,tang2023mvdiffusion,xiang20233d,liu2023deceptive,lei2022generative,zhang2023text2nerf} are made to extend 2D diffusion models from single-view images to multi-view consistent images.
However, most of these methods focus on image generation and are not designed for 3D reconstruction.
The recent works Viewset Diffusion~\cite{szymanowicz2023viewset}, SyncDreamer~\cite{liu2023syncdreamer}, and MVDream~\cite{shi2023mvdream} share a similar idea to produce consistent multi-view color images via attention layers. However, 
unlike that normal maps explicitly encode geometric information, reconstruction from color images always suffers from texture ambiguity, and, thus, they either struggle to recover geometric details or require huge computational costs. For instance, SyncDreamer~\cite{liu2023syncdreamer} requires dense views for 3D reconstruction but still suffers from low-quality geometry and blurring textures.
MVDream~\cite{shi2023mvdream} still resorts to a time-consuming optimization using SDS loss for 3D reconstruction, and its multi-view distillation scheme requires 1.5 hours. 

Direct2.5\cite{lu2024direct2} employs a sequential process: it first generates multi-view normal maps and subsequently generates RGB images. Consequently, the generation of normal maps and RGB images is largely decoupled. This separation can be less conducive to achieving strong alignment between the geometric and textural domains, or to effectively learning a coherent 3D data distribution. Moreover, the inconsistency between the camera type of the input images and the predefined camera type leads to unnatural distortions in the generated meshes.
In contrast, thanks to our proposed domain switcher, our method can simultaneously generate multi-view normal maps for geometric reconstruction and RGB images for coloring. Additionally, our method provides the flexibility to generate results in either orthogonal or perspective camera settings, adapting to different requirements as needed.
 
\subsection{3D Generative Models}
Instead of performing a time-consuming per-shape optimization guided by 2D diffusion models, some works attempt to directly train 3D diffusion models based on various 3D representations, like point clouds~\cite{nichol2022point, zeng2022lion, luo2021diffusion, zhou20213d}, meshes~\cite{liu2023meshdiffusion,gao2022get3d}, neural fields~\cite{wang2023rodin,cheng2023sdfusion,karnewar2023holofusion, kim2023neuralfield, gu2023learning, anciukevivcius2023renderdiffusion, muller2023diffrf, ntavelis2023autodecoding, jun2023shap, zhang20233dshape2vecset, gupta20233dgen, erkocc2023hyperdiffusion, chen2023single}.
However, due to the limited size of publicly available 3D assets datasets, most of the works have only been validated on limited categories of shapes or lack of sufficient generalization properties, and how to scale up on large datasets is still an open problem. 
In order to effectively improve the generalization performance of the model, a portion of recent work~\cite{wang2024crm, xu2024instantmesh, tang2024lgm, xu2024grm, li2024craftsman} seeks to first utilize the multi-view diffusion model to infer multi-view images and then use sparse view reconstruction models to obtain meshes in a single forward pass. 
Such sparse view reconstruction models are usually based on transformer architectures to predict Nerf~\cite{xu2024instantmesh}, Gaussian Splatting~\cite{tang2024lgm, xu2024grm} or triplane~\cite{wang2024crm} representations and employ differentiable rendering to compute the loss function, which hinders the efficiency and stability of the training process. 
As a result, the training process typically requires tens or hundreds of GPUs and it is difficult to obtain geometries and textures with sufficient details.
On the contrary, our method first adopts 2D representations, enabling it to leverage 2D diffusion models~\cite{rombach2022high}, whose pre-trained priors significantly enhance zero-shot generalization ability. 
Building on this, we then employ a cascaded approach to efficiently reconstruct 3D meshes, improving both geometric and texture quality while avoiding the high computational costs associated with training a feedforward reconstruction model.

%% file: sections/3_method.tex
\section{Problem Formulation}

\subsection{Diffusion Models} 
Diffusion models~\cite{ho2020denoising,sohl2015deep} are first proposed to gradually recover images from a specifically designed degradation process, where a forward Markov chain and a Reverse Markov chain are adopted. 
Given a sample $z_0$ drawn from the data distribution $p(z)$, the forward process of denoising diffusion models yields a sequence of noised data $\left\{z_t \mid t \in(0, T)\right\}$ with $z_t=M_t z_0+\sigma_t \epsilon$, where $\epsilon$ is random noise drawn from distribution $\mathcal{N}(0,1)$, and $M_t, \sigma_t$ are fixed sequence of the noise schedule. 
The forward process will be iteratively applied to the target image $z$ until the image becomes complete Gaussian noise at the end. 
On the contrary, the reverse chain iteratively denoises the corrupted image, recovering $z_{t-1}$ from $z_t$ by step-by-step predicting the added random noise $\epsilon$.

\subsection{The Distribution of 3D Assets} 
\label{3d_distribution}

Unlike prior works that adopt 3D representations like point clouds, tri-planes, or neural radiance fields, we propose that the distribution of 3D assets denoted
as $p_a(\mathbf{z})$, can be modeled as a joint distribution of its
corresponding 2D multi-view normal maps and color images.  Specifically, given a set of cameras $\left\{\boldsymbol{\pi}_1, \boldsymbol{\pi}_2, \cdots, \boldsymbol{\pi}_K\right\}$ and a conditional input image $y$, we have
\begin{equation}
\label{eqn:joint_distribution}
p_a(\mathbf{z})=p_{nc}\left(N^{1:K}, X^{1:K} | y \right),
\end{equation}
where $p_{nc}$ is the distribution of the normal maps $N^{1:K}$ and color images $X^{1:K}$ observed from a 3D
asset conditioned on an image $y$. 
For simplicity, we omit the symbol $y$ for this equation in the following discussions.
Therefore, our goal is to learn a model $f$ that synthesizes multiple normal maps and color images of a set of camera poses denoted as
\begin{equation}
\label{eqn: model_func_0}
(N^{1:K}, X^{1:K})=f(y, \boldsymbol{\pi}_{1:K}).
\end{equation}

Finally, we can formulate this cross-domain joint distribution as a Markov chain within the diffusion scheme:

\begin{equation}
\begin{split}
    & p_{nc}(X_{1:T},N_{1:T}) = p\left({N}^{(1:K)}_T, {X}^{(1:K)}_T\right) \\
    &\quad \times \prod_t p_{nc}\left({N}^{(1:K)}_{t-1}, {X}^{(1:K)}_{t-1} \;\middle|\; {N}^{(1:K)}_{t}, {X}^{(1:K)}_{t}\right),
\end{split}
\end{equation}
where $p\left({N}^{(1:K)}_T, {X}^{(1:K)}_T\right)$ are random Gaussian noises. Our key problem is to characterize the joint distribution $p_{nc}$, so that we can sample from this Markov chain to generate normal maps and images.

\subsection{Mesh Reconstruction via Differentiable Rendering}
\label{sec:remeshing method} 
Given the synthesized multi-view normal maps $N^{1:K}$, we extract the 3D mesh using differentiable rendering, ensuring accurate geometry reconstruction through gradient-based optimization. Specifically, starting with an initial triangle mesh $M$, composed of vertices $V$ and faces $F$, the multi-view normal maps $N^{i:K}$ and masks $M^{i:K}$ are rendered differentiably using the nvdiffrast library~\cite{laine2020modular}. The reconstruction loss is then computed, and the gradient is back-propagated to optimize the vertices $V$, refining the mesh geometry based on the multi-view input.

To directly optimize the vertices of the mesh and continuously adjust the topology of the mesh and the edge lengths of the mesh, we adopt the remeshing strategy proposed by ~\cite{palfinger2022continuous}, and apply it after each optimization iteration. This process involves edge-splitting, edge-collapsing, edge-flipping, and smoothing to adaptively adjust local edge lengths and correct defects caused by sparse or inconsistent vertex gradients. 

The reasons we choose an explicit mesh representation over implicit methods like Neus~\cite{wang2021neus} are twofold: 1) The \textbf{computational complexity} of optimizing with the explicit method scales with the number of mesh vertices and the square of the resolution of the multi-view normal maps, whereas in implicit neural field methods, it scales with the cube of the image resolution, making our approach more efficient. 2) Directly optimizing mesh vertices and faces allows our method to \textbf{integrate seamlessly} with the cascaded mesh extraction process, eliminating the computational overhead and detail loss that typically occur during conversions between mesh and implicit SDF representations.

\section{Method}
As per our problem formulation in Section~\ref{3d_distribution}, we propose a multi-view cross-domain diffusion scheme, which operates on two distinct domains to generate multi-view consistent normal maps and color images. The overview of our method is presented in Figure~\ref{fig:pipeline}. 
First, our method adopts a \textbf{ cross-domain multi-view diffusion} scheme to generate six views of normal maps and color images and enforces consistency across different views using multi-view attention.
Second, our proposed domain switcher allows the diffusion model to operate on more than one domain.
A \textbf{cross-domain attention} is proposed to propagate information between the normal domain and color image domain ensuring geometric and visual coherence between the two domains (see Section~\ref{sec:cross-domain}).
Furthermore, to achieve high-quality geometry and appearance from multi-view 2D normal and color images while minimizing blurring from sparse views and multi-view inconsistencies, we introduce a \textbf{cascaded 3D mesh extraction algorithm}. This algorithm reconstructs the 3D mesh in a coarse-to-fine manner, iteratively refining it with our cross-domain multi-view enhancement module (see Section~\ref{sec:mesh-extraction}).

\subsection{Multi-view Cross-domain Diffusion Model}
\label{sec:cross-domain}
Our model is built upon the pre-trained stable diffusion models~\cite{radford2021learning} to leverage its strong priors. 
Given that current 2D diffusion models~\cite{radford2021learning,liu2023zero} are designed for a single domain, the main challenge lies in how to effectively extend stable diffusion models to support multi-domain operations. To achieve this goal, we explore several possible designs.

\subsubsection{Naive Solutions for Cross-domain Generation}
Building upon a pre-trained diffusion model with a UNet backbone, a straightforward solution is to add four more channels to the output of the UNet module representing the extra domain. Therefore, the diffusion model can simultaneously output normals and color image domains. However, we notice that such a design suffers from low convergence speed and poor generalization. This is because the channel expansion may perturb the pre-trained weights of stable diffusion models and therefore cause catastrophic model forgetting. In contrast, our method, leveraging the domain switcher introduced later, circumvents this issue by not requiring any modification to the number of channels in the pre-trained model.

Revisiting Eq.~\ref{eqn:joint_distribution}, it is possible to factor the joint distribution into two conditional distributions:
\begin{equation}
\label{eqn:conditional_distribution}
q_a(\mathbf{z})=  q_n (N^{1:K}) \cdot  q_c\left(X^{1:K} \mid N^{1:K}\right) .
\end{equation}
This equation suggests an alternative solution where we could initially train a diffusion model to generate normal maps and then train another diffusion model to produce color images, conditioning on the generated normal maps (or vice versa). 

However, implementing this two-stage framework introduces certain complexities and significantly increases computational costs. More importantly, as discussed in Section~\ref{3d_distribution}, multi-view color images and normal maps share a joint distribution in 3D data, where each domain influences the other. Jointly learning these domains facilitates cross-domain corroboration, enabling each domain to benefit from the other. In contrast, a sequential learning strategy trains the two stages separately, resulting in a larger domain gap and performance degradation. This occurs because the second-stage model is trained on ideal GT images but must infer from the imperfect images predicted by the first-stage model. For a detailed discussion, please refer to Section \ref{sec:discussion}. 

\subsubsection{Domain Switcher} To overcome these difficulties mentioned above and model the joint distribution of the multi-view color images and normal maps, we design a cross-domain diffusion scheme via a \textbf{\textit{domain switcher}}, denoted as $s_d$.
The switcher $s_d$ is a one-dimensional vector serving as labels for distinct domains, and we further feed the switcher into the diffusion model as an extra input.
Therefore, the formulation of Eq.~\ref{eqn: model_func_0} can be extended as:
\begin{equation}
\begin{aligned}
    N^{1:K}, X^{1:K}=f(y, \boldsymbol{\pi}_{1:K}, s_{d} = n), f(y, \boldsymbol{\pi}_{1:K}, s_{d} = c).
\end{aligned}
\label{eqn: model_func_1}
\end{equation}

The domain switcher is first encoded via positional encoding~\cite{mildenhall2020nerf} and subsequently concatenated with the time embedding. This combined representation is then injected into the UNet of the stable diffusion models. Interestingly, experiments show that this subtle modification does not significantly alter the pre-trained priors. As a result, it allows for fast convergence and robust generalization, without requiring substantial changes to the stable diffusion models.

\input{figures/trans-block}

\subsubsection{Cross-domain Attention} 
Using the proposed domain switcher, the diffusion model can generate two different domains. However, it is important to note that for a single view, there is no guarantee that the generated color image and the normal map will be geometrically consistent if the color images and normal maps are sampled independently from the distribution.
To address this issue and ensure consistency between the generated normal maps and color images, we introduce a cross-domain attention mechanism to facilitate the exchange of information between the two domains. This mechanism aims to ensure that the generated outputs align well in terms of geometry and appearance.

Given a latent embedding $Z$, the original self-attention mechanism~\cite{vaswani2017attention} can be defined as:
\begin{equation}
\begin{aligned}
Attn(Q, K, V) = Softmax(\frac{QK^T}{\sqrt{d}})\cdot V
\end{aligned}
\label{eqn: self-attn}
\end{equation}
where the query, key, and value are projected by learnable matrices as follows:
\begin{equation}
\begin{aligned}
    Q = W^Q Z,\quad K = W^K Z,\quad V = W^V Z
\end{aligned}
\label{eqn: QKV}
\end{equation}
\noindent
Our cross-domain attention layer maintains a similar structure as the original self-attention layer and is integrated seamlessly before the cross-attention layer in each transformer block of the UNet, as depicted in Figure~\ref{fig:trans_block}.
In the cross-domain attention layer, given the $Z_n$ and $Z_x$ represent the latent feature of the normal domain and the color domain respectively, $[\cdot]$ signifies concatenation operation, the key and value are derived as follows:
\begin{equation}
\begin{aligned}
    K = W^K [Z_n, Z_x],\quad V = W^V [Z_n, Z_x]
\end{aligned}
\label{eqn: model_func_1}
\end{equation}
\noindent
We then apply attention operation specifically as:
\begin{equation}
\begin{aligned}
    Z_n = Attn(Z_n, K, V), \quad Z_x = Attn(Z_x, K, V)
\end{aligned}
\label{eqn: model_func_1}
\end{equation}
\noindent
This design ensures that the generations of color images and normal maps are closely correlated, thus promoting geometric consistency between the two domains.

\subsubsection{Multi-view Cross-domain Transformer Block} 
The prior 2D diffusion models~\cite{radford2021learning,liu2023zero} generate each image separately, so that the resulting images are not geometrically and visually consistent across different views. 
To enhance consistency among different views, similar to prior works such as SyncDreamer~\cite{liu2023syncdreamer} and MVDream~\cite{shi2023mvdream}, we utilize attention mechanism to facilitate information propagation across different views, implicitly encoding multi-view dependencies (as illustrated in Figure~\ref{fig:trans_block}). This is achieved by extending the original self-attention layers to be global-aware, allowing connections to other views within the attention layers. Keys and values from different views are connected to each other to facilitate the exchange of information. By sharing information across different views and domains within the attention layers, the model perceives cross-domain multi-view correlation and is capable of generating consistent multi-view color images and normal maps.

\subsubsection{Camera Type Switcher} 
The input images typically come from diverse sources, such as real photographs, artist-created figures, or images generated by 2D AIGC tools, resulting in significant variation in their projection types. For instance, camera-captured images usually follow perspective projection, while synthetic images are often better suited to orthogonal projection. Ignoring these underlying camera projections makes it challenging to generate multi-view consistent and visually plausible images, leading to distortions in the reconstructed mesh.
To handle varying camera settings in input images, we introduce a one-dimensional camera type switcher, \( s_c \), allowing users to specify the projection type, such as orthogonal or perspective. For perspective projection, we adopt the standard 35mm focal length. The camera switcher is integrated into the model similarly to the domain switcher, enabling the generation of multi-view images consistent with the specified camera settings.

\subsubsection{Multi-Stage Training Scheme}
To enable a seamless transition from a pre-trained stable diffusion model to the target joint distribution of 3D objects, we propose a multi-stage training strategy consisting of three stages.

\paragraph{Multi-Domain Pre-Training Stage} Building on the stable diffusion model, we initially remove the domain switcher and cross-domain attention layers, modifying only the self-attention layers into a multi-view design. Here, the model is fine-tuned for multi-view generation from a single view within the same domain, predicting multi-view colors, normals, or masks from a single view of color, normal, or mask, respectively. This stage equips the model with robust multi-view prior knowledge across various domains.

\paragraph{Mixed-Domain Fine-Tuning Stage} In this stage, we introduce the domain switcher, fine-tuning the model to generate either multi-view colors or normals from a single-view color image, guided by the domain switcher. This enables the model to produce multi-view outputs in either the color or normal domain from a single-view color image.

\paragraph{Cross-Domain Alignment Fine-Tuning Stage} In the final stage, we incorporate cross-domain attention layers while freezing all previously trained parameters. The model is then fine-tuned to jointly predict both multi-view color images and normal maps from a single-view color image. The cross-domain attention layers facilitate information exchange across color and normal domains, enhancing the consistency between generated normal maps and color images.

This multi-stage training scheme enhances training stability and leverages multi-task learning principles, enabling our multi-view cross-domain diffusion model to generate high-quality multi-view color images and normal maps with excellent generalizability.

\subsection{Cascaded 3D Mesh Extraction}
\label{sec:mesh-extraction}
To efficiently extract explicit 3D geometry from 2D normal maps and color images, we use the mesh-based differentiable rendering algorithm described in Section~\ref{sec:remeshing method}, optimizing directly on the mesh vertices $V$ guided by the generated 2D representations. 
Our cascaded 3D mesh extraction process is divided into three stages:
1) \textit{Geometric Initialization}: Initialize a rough mesh with approximate topology for optimization.
2) \textit{Inconsistency-Aware Coarse Reconstruction}: Reconstruct a coarse shape and texture using the generated multi-view color images and normals.
3) \textit{Iterative Refinement}: Iteratively refine the geometry and texture through a cross-domain enhancement module.

\subsubsection{Geometric Initialization}
Explicit mesh optimization methods typically require a mesh with approximate topology for initialization. Following~\cite{wu2024unique3d}, we first use normal map integration from the front and back views to obtain two corresponding depth maps. These depth maps are then projected into 3D space to create two point cloud parts with normal, which are combined into an initialized mesh using the Poisson reconstruction~\cite{kazhdan2006poisson} method. Unlike the commonly used sphere initialization, this approach not only speeds up optimization but also enables our method to capture relatively complex shapes by providing an approximate topological structure.

However, this method does not always yield initializations with the right topology. As shown in Figure~\ref{fig:init}, when the target object’s geometry is concave rather than convex, this strategy leads to failed initialization. This issue arises because the depth recovered from a monocular normal map is scale ambiguous and often shifted relative to the true depth. Consequently, the mesh parts may be incorrectly combined, with significant overlap, due to inaccurate depth information, resulting in initializations with completely wrong topology.
To address this, we introduce a simple yet effective concave topology-checking strategy. Specifically, for each depth map (front and back views), we calculate the average depth of the entire map (\(d_2\)) and of a central sub-region (\(d_1\)). If the front of the object is concave, \(d_1\) should be greater than \(d_2\) (i.e., \(d_1 > d_2\)) for the front view.  Similarly, if the back of the object is concave, \(d_2\) will be greater than \(d_1\) for the back view.  If either condition is met, the object is likely concave, and we apply sphere initialization instead. This strategy effectively enhances the robustness of the initialization process.

\input{figures/initialization_method}

\subsubsection{Coarse Reconstruction} 
\label{sub:coarse_recon}
To extract explicit 3D geometry from 2D normal maps and color images, we optimize an explicit mesh to amalgamate all 2D generated data. 
Nonetheless, adopting existing differentiable rendering-based optimization schemes proves unviable. These methods were tailored for real-captured images and necessitate dense input views. In contrast, our generated views are relatively sparse, and the generated normal maps and color images may exhibit subtle inaccurate predictions of some pixels. Regrettably, these errors accumulate during the geometry optimization, leading to distorted geometries, outliers, and incompleteness.
To overcome the challenges above, we propose a novel geometric-aware optimization scheme. 

\paragraph{Geometric Carving}
With the initialized mesh, we leverage differentiable rendering technique to optimize 3D shapes using the generated multi-view normals. Specifically, we render the normal maps $N_{r}^{1:K}$ and masks $M_{r}^{1:K}$ via a differentiable renderer, and then minimize the differences between the rendered ones and generated ones.  The objective loss function is formulated as follows:
\begin{equation}
\begin{aligned}
\mathcal{L} &= \mathcal{L}_{normal} + \mathcal{L}_{mask} + L_{geo} + \mathcal{R}_{Laplace}
\end{aligned}
\label{eqn: recon_loss}
\end{equation}
where $\mathcal{L}_{normal}$ denotes the $L_2$ loss between the rendered normals and the generated normals, $\mathcal{L}_{mask}$ denotes $L_2$ loss between the rendered masks and the masks segmented from the generated normals, $L_{geo}$ denotes the geometry-aware normal loss which we will discuss later, and $\mathcal{R}_{Laplace}$ denotes a Laplace regularization term that calculates the position deviation of a vertex with its neighbors to encourage the smoothness of the mesh surface.

Through the initialization and geometric carving process mentioned above, we obtain a mesh with an approximately correct shape. However, due to the presence of several heuristic operations during the remeshing step and the challenge of ensuring perfect multi-view consistency when generating normal maps, the optimized mesh still lacks fine geometric details and smoothness.

\input{figures/color-compare}

\paragraph{Geometry-aware Normal Loss}
Thanks to the differentiable rendering technique, we can easily extract the surface normals $\hat{\mathbf{g}}$ of the optimized mesh by leveraging gradients of the surface with respect to its spatial coordinates.
We maximize the similarity of the normal of the mesh $\hat{\mathbf{g}}$ and our generated normal $\mathbf{g}$ to provide 3D geometric supervision.
For a 3D point $p$ on the optimized mesh, it will be visible by multiple viewpoints but its corresponding normals of the multiple viewpoints $\mathbf{g}_p^{1:K}$ may have subtle variance, thus leading to trivial inaccuracies. To tolerate trivial inaccuracies of the generated normals from different views, we introduce a geometry-aware normal loss. 

\begin{equation}
\mathcal{L}_{geo} = \sum_{p\in \mathcal{P}}  w_p^{i} \cdot | \hat{\mathbf{g}}_p - \mathbf{g}_p^{i}  |
\end{equation}
where
\begin{equation}
\label{eq: geo_confidence}
\centering
w_p^{i} =\frac{\bar{w}_p^{i}}{\sum_{i=1}^{K} \bar{w}_p^{i}} 
\end{equation}

\begin{equation}
\centering
    \bar{w}_p^{i} = \begin{cases}0, & \cos \left(\mathbf{v}_k, \mathbf{g}_k\right)>\epsilon \\
m_p^{i} \cdot | \cos \left(\mathbf{v}_p^{i}, \mathbf{\hat{g}}_p^{i}\right) | , & \cos \left(\mathbf{v}_p^{i}, \mathbf{g}_p^{i}\right) \leq \epsilon\end{cases} 
\end{equation}
Here $|\cdot|$ denotes an absolute function, $m_p^{i}$ is the visible mask and $\epsilon$ is a negative threshold closing to zero. For a 3D point $p$ located in the optimized mesh, we measure the cosine value of the angle between its current normal $g_k$ and its corresponding ray's viewing direction $\mathbf{v}_k$ across the generated multiple viewpoints. 

The design rationale behind this approach lies in the orientation of normals, which are deliberately set to face outward, while the viewing direction is inward-facing. This configuration ensures that the angle between the normal vector and the viewing ray remains not less than $90^{\circ}$. A deviation from this criterion would imply inaccuracies in the generated normals.
Furthermore, rather than treating normals from different views equally, we introduce a weighting mechanism. We assign higher weights to normals that its corresponding viewing ray direction maintains larger angle with the current mesh normal. This prioritization enhances the accuracy of our geometric supervision process.

\input{figures/gso_compare}
\input{tables/reconstruction} 

\paragraph{UV-based Texture Generation}
The prior works store texture information in mesh vertices, so that the texture quality is limited by the density of mesh vertices. In contrast, we adopt UV map to store texture information to maintain high-quality texture without information loss. Specifically, a simple yet efficient two-stage strategy is used to create a high-quality UV texture map with enhanced integrity and coherence from multi-view color images generated by the cross-domain diffusion model while mitigating the texture detail loss due to optimization.

In the first stage, we derive the vertex colors of the mesh from multi-view color images. 
For a vertex $v$ on the mesh, it may be visible by several neighboring viewpoints, each corresponding to a multiple pixel value $x^i$ in camera spaces with slight variance.
The vertex color $c$ is computed by fusing these multiple pixel values as $c=\sum{w_p^i  x^i}$.
Rather than treating the color information of all viewpoints equally, such a vertex texturing process takes into account the imperfect consistency of the pixel level of multi-view images by an adaptive weighting strategy. 
Due to the limited viewpoints and self-occlusion of the object, some vertices remain unobserved by any view (i.e., invisible vertices), resulting in texture gaps. 
To address this, starting from the well-colored vertices, we gradually propagate color information to adjacent uncolored vertices.
During each propagation step, the color attribute of each uncolored vertex is updated with the average color of its neighboring vertices, yielding a continuous and complete texture across the meshing.

In the second stage, we unwrap the color of the vertices into UV space, yielding an initial UV texture map. 
Next, we project the multi-view images onto the UV space to create a UV texture, which is then blended with the initial texture to enhance and refine its details.
Finally, dilation operations are operated to fill in small uncolored regions. This strategy preserves the overall texture coherence achieved through vertex colorization, while also enhancing finer details in UV space, resulting in a more accurate and visually cohesive texture across the mesh surface.

\subsubsection{Iterative Refinement}
With the proposed operations above, a textured mesh is recovered from the generated multi-view color images and normal maps. Due to the mesh still containing imperfections, we name it coarse mesh and would like to further improve it.
We leverage a cross-domain multi-view enhancement module to provide high-resolution 2D guidance for optimization.
Specifically, as shown in Fig.~\ref{fig:pipeline}, we render 4 views of RGB images and normal maps from the coarse mesh, and subsequently the rendered images and normals are enhanced via the cross-domain multi-view enhancement module to produce high-quality images and normals with a $512 \times 512$ resolution.
The enhanced high-resolution images and normals are then used to improve the coarse mesh via differentiable rendering.

\input{figures/condx.tex}

\paragraph{Cross-domain Multi-view Enhancement Module} The module uses the ControlNet~\cite{zhang2023adding} architecture to fully leverage prior information from stable diffusion. We treat the ControlNet as a trainable module while keeping the pre-trained stable diffusion weights fixed. To maintain consistency between the enhanced normal maps and color images, we integrate cross-domain attention into the ControlNet. Low-quality RGB images and normal maps are input into the enhancement module, producing corresponding high-quality outputs with 2x resolution upsampling and enhancement. Additionally, the original input image is processed through an IP-Adapter~\cite{ye2023ip} to incorporate high-level textual information.

\input{tables/nvs}

\paragraph{Iterative High-resolution Refinement} 
We first render normal maps and color images of the currently optimized mesh, feeding these into the cross-domain multi-view enhancement module to produce enhanced outputs. Unlike in the training stage, the low-resolution color images and normal maps generated in Sec.~\ref{sec:cross-domain} are used as conditioning inputs for the ControlNet branch, while Gaussian noise is added to the rendered outputs, which are then fed into the stable diffusion branch for denoising. 
This process, known as DDIM inversion, establishes a robust starting point for the iterative denoising process, enabling the refinement of surface imperfections and blurring on the existing mesh.
With these enhanced multi-view results, we iteratively refine geometry and texture through the proposed reconstruction scheme (see Sec.~\ref{sub:coarse_recon}). Ultimately, this approach produces a high-quality textured mesh with intricate geometric and visual details.

%% file: figures/trans-block.tex
\begin{figure}[tp!]
\centering
\includegraphics[width=0.9\linewidth]{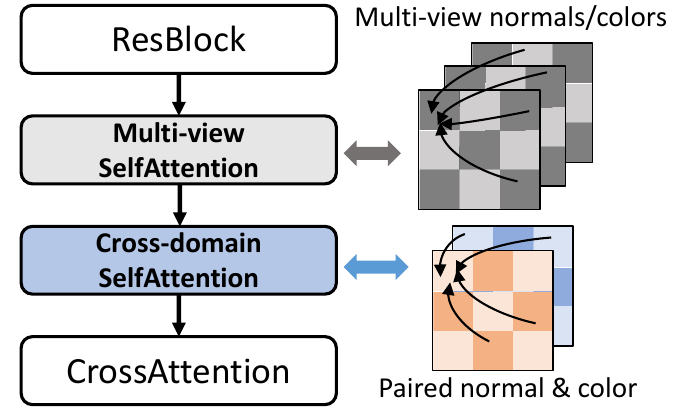}
\caption{The illustration of the structure of the multi-view cross-domain transformer block, where the multi-view attention layer and cross-domain attention layer facilitate information exchange across different views and domains, respectively.}
\vspace{-2mm}
\label{fig:trans_block}	
\end{figure}

%% file: figures/initialization_method.tex
\begin{figure}[tp!]
\centering
\includegraphics[width=1\linewidth]{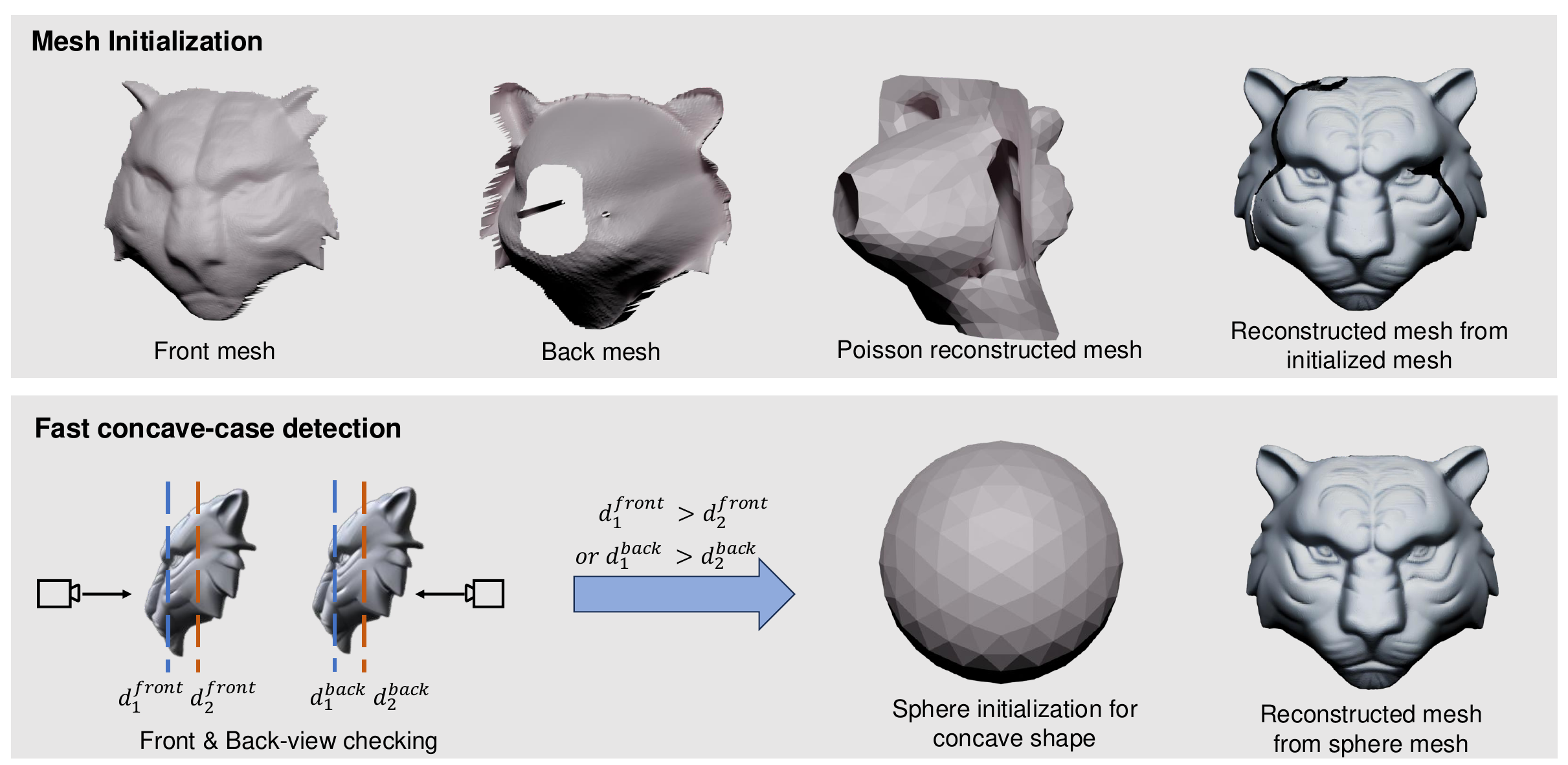}
\caption{The illustration of our geometric initialization strategy. We employ the Poisson reconstruction~\cite{kazhdan2006poisson} method for geometric initialization using normal maps from front and back views, while estimated depth maps are used to detect and correct potential initialization errors.}
\vspace{-2mm}
\label{fig:init}	
\end{figure}

%% file: figures/color-compare.tex
\begin{figure*}[htp!]
\centering
\includegraphics[width=\linewidth]{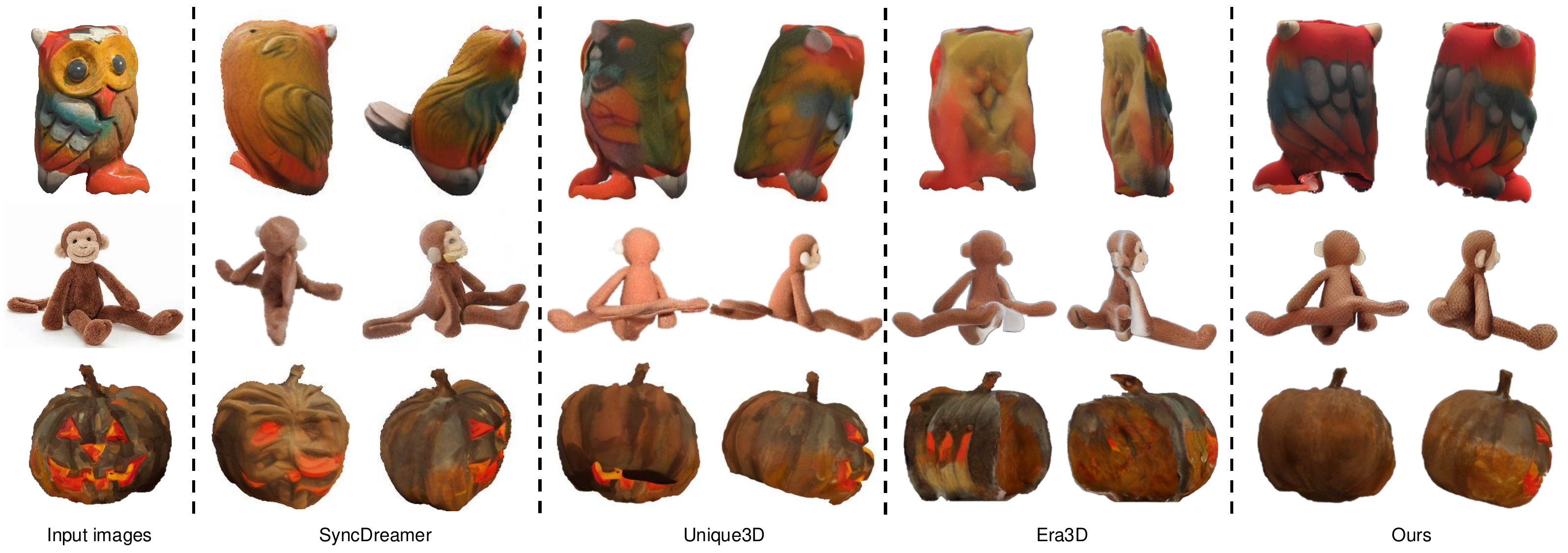}
\caption{The qualitative comparisons with baseline models on synthesized multi-view color images.}
\vspace{-3mm}
\label{color-compare}	
\end{figure*}

%% file: figures/gso_compare.tex
\begin{figure*}[t]
\centering
\includegraphics[width=\linewidth]{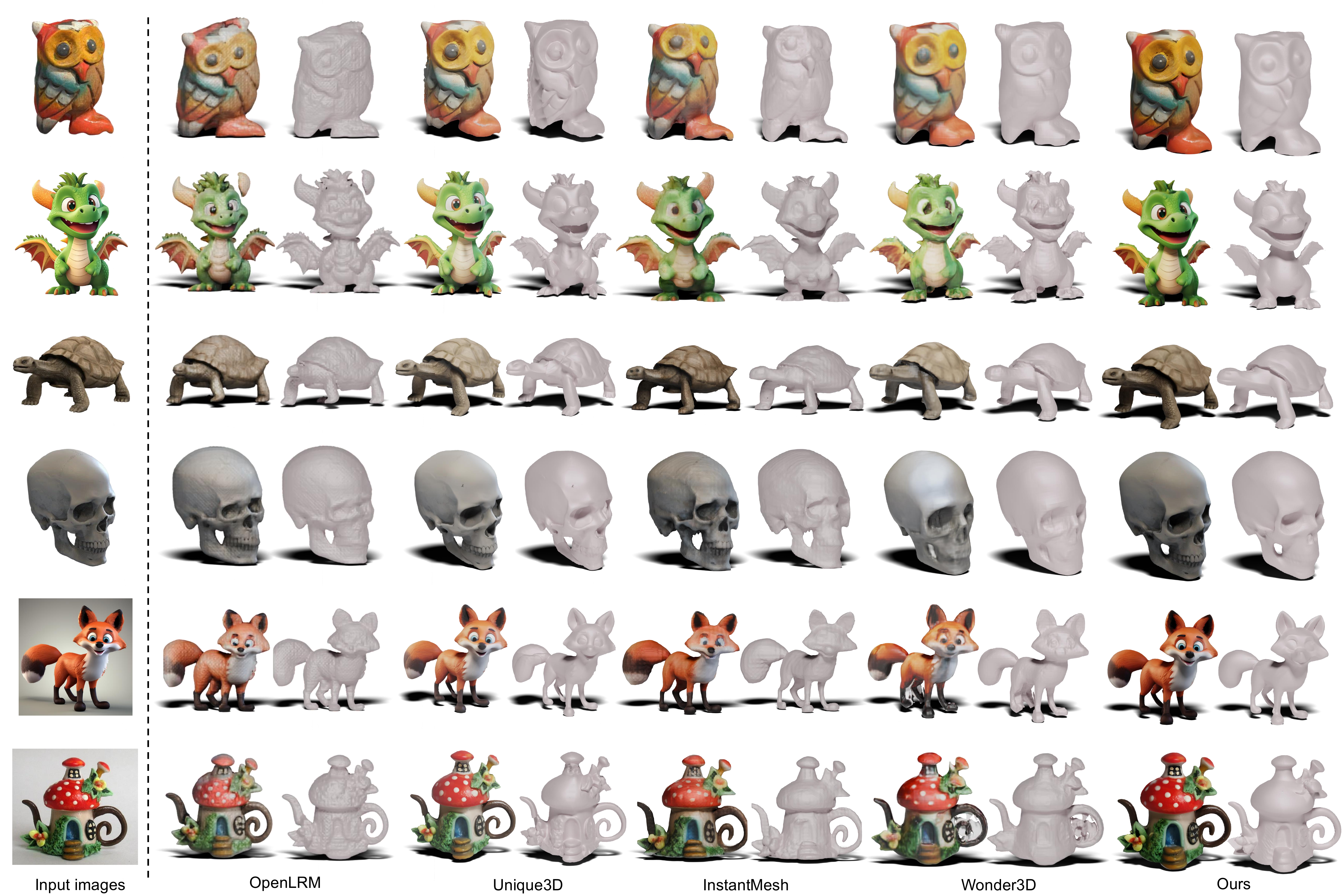}
\caption{The qualitative comparisons with baseline methods on in-the-wild images in terms of the reconstructed textured meshes.}
\vspace{-5mm}
\label{gso_compare}	
\end{figure*}

%% file: tables/reconstruction.tex
\begin{table}[t]
    \caption{Quantitative comparison with baselines on the single-view reconstruction task. We report Chamfer Distance and Volume IoU on the GSO dataset for both orthogonal and perspective settings, where $\pm$ indicates the standard deviation.}
    \centering
    \resizebox{0.85\linewidth}{!}{
    \begin{tabular}{lcc}
        \toprule
        Method  & Chamfer Dist. $\downarrow$ \textcolor{gray}{(STD)}  & Volume IoU $\uparrow$ \textcolor{gray}{(STD)}  \\
        \midrule
        Realfusion~\cite{melas2023realfusion}   
        & 0.0819 \textcolor{gray}{\textsuperscript{$\pm$ 0.0010}}  & 0.2741 \textcolor{gray}{\textsuperscript{$\pm$ 0.0092}}  \\
        Magic123~\cite{qian2023magic123}
        & 0.0516 \textcolor{gray}{\textsuperscript{$\pm$ 0.0021}} &  0.4528 \textcolor{gray}{\textsuperscript{$\pm$ 0.0183}} \\
        One-2-3-45~\cite{liu2023one}   
        & 0.0629 \textcolor{gray}{\textsuperscript{$\pm$ 0.0009}} &  0.4086 \textcolor{gray}{\textsuperscript{$\pm$ 0.0125}} \\
        Point-E~\cite{nichol2022point}   
        & 0.0426 \textcolor{gray}{\textsuperscript{$\pm$ 0.0011}} & 0.2875 \textcolor{gray}{\textsuperscript{$\pm$ 0.0071}} \\
        Shap-E~\cite{jun2023shap}   
        & 0.0436 \textcolor{gray}{\textsuperscript{$\pm$ 0.0018}} &  0.3584 \textcolor{gray}{\textsuperscript{$\pm$ 0.0088}} \\
        Zero123~\cite{liu2023zero}   
        & 0.0339 \textcolor{gray}{\textsuperscript{$\pm$ 0.0015}} &  0.5035 \textcolor{gray}{\textsuperscript{$\pm$ 0.0084}} \\
        SyncDreamer~\cite{liu2023syncdreamer}  
        &  0.0261 \textcolor{gray}{\textsuperscript{$\pm$ 0.0062}}  &  0.5421 \textcolor{gray}{\textsuperscript{$\pm$ 0.0079}}   \\
        OpenLRM~\cite{hong2023lrm}
        & 0.0255 \textcolor{gray}{\textsuperscript{$\pm$ 0.0027}}  &  0.5452 \textcolor{gray}{\textsuperscript{$\pm$ 0.0064}} \\
        InstantMesh~\cite{xu2024instantmesh} 
        &  0.0224 \textcolor{gray}{\textsuperscript{$\pm$ 0.0015}}  & 0.5353 \textcolor{gray}{\textsuperscript{$\pm$ 0.0084}}  \\
        CRM~\cite{wang2024crm}
        &  0.0220 \textcolor{gray}{\textsuperscript{$\pm$ 0.0009}}   &  0.5412 \textcolor{gray}{\textsuperscript{$\pm$ 0.0059}}   \\
        Unique3D~\cite{wang2024crm}
        &  0.0238 \textcolor{gray}{\textsuperscript{$\pm$ 0.0023}}  &  0.5134 \textcolor{gray}{\textsuperscript{$\pm$ 0.0112}}   \\
        Era3D ~\cite{li2024era3d}
        &  0.0239 \textcolor{gray}{\textsuperscript{$\pm$ 0.0026}}  &  0.5340 \textcolor{gray}{\textsuperscript{$\pm$ 0.0075}}  \\
        Wonder3D ~\cite{long2024wonder3d}    
        &  0.0222 \textcolor{gray}{\textsuperscript{$\pm$ 0.0011}} &  0.5521 \textcolor{gray}{\textsuperscript{$\pm$ 0.0066}}   \\
        Ours(Perspective Setting)
        &  0.0206 \textcolor{gray}{\textsuperscript{$\pm$ 0.0015}}  &  {0.6326 \textcolor{gray}{\textsuperscript{$\pm$ 0.0094}}}   \\
        Ours(Orthogonal Setting)
        &  \textbf{0.0193 \textcolor{gray}{\textsuperscript{$\pm$ 0.0013}}}  &  \textbf{0.6402 \textcolor{gray}{\textsuperscript{$\pm$ 0.0081}}}   \\
        \bottomrule
    \end{tabular}
    }
    \label{tab:recon}
\vspace{-4mm}
\end{table}

%% file: figures/condx.tex
\begin{figure*}[t]
\centering
\includegraphics[width=\linewidth]{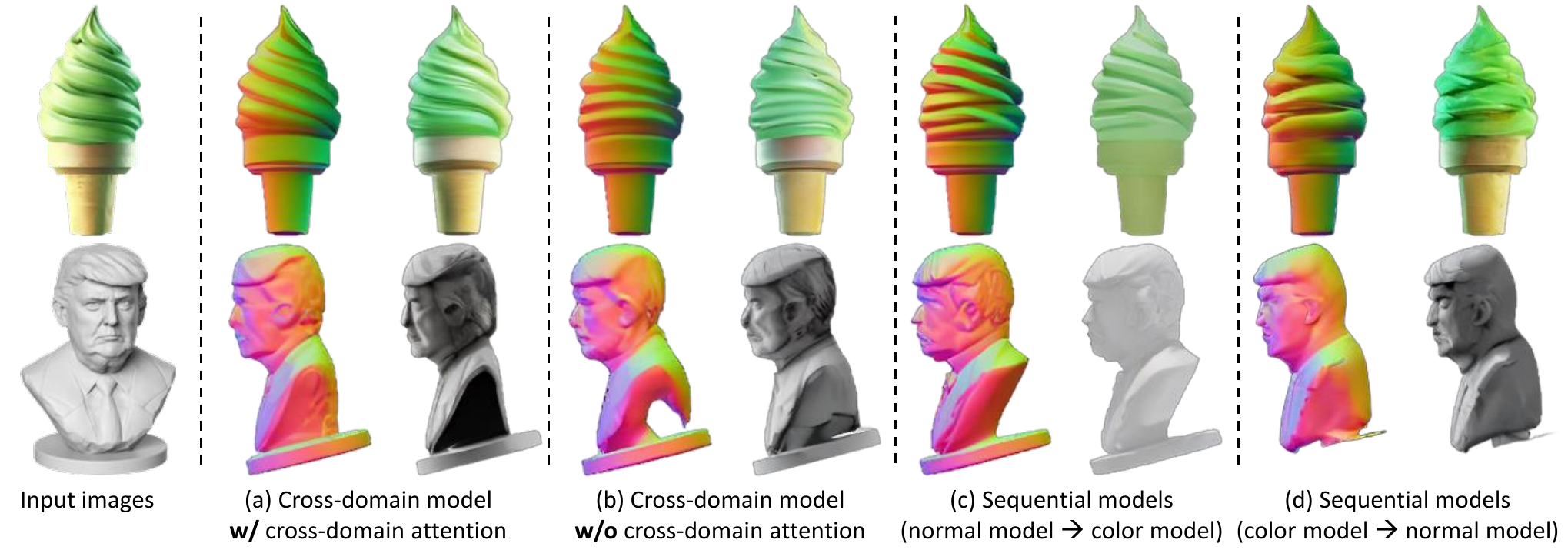}
\caption{Ablation studies on different cross-domain diffusion schemes. Jointly predicting multi-view normal maps and color images provides a more integrated perception of information across domains. Additionally, the proposed cross-domain attention mechanism enhances the consistency between the generated normal maps and color images, ensuring improved alignment and quality in the outputs.
}
\vspace{-4mm}
\label{fig:ablation_condx}	
\end{figure*}

%% file: tables/nvs.tex
\begin{table}[t]
    \caption{{Quantitative comparison for novel view synthesis. We report PSNR, SSIM~\cite{wang2004image}, LPIPS~\cite{zhang2018unreasonable} on the GSO~\cite{downs2022google} dataset, where $\pm$ indicates the standard deviation.}}
    \centering
    \resizebox{0.75\linewidth}{!}{
    \begin{tabular}{lccc}
        \toprule
        Method & PSNR $\uparrow$ \textcolor{gray}{(STD)}  & SSIM $\uparrow$ \textcolor{gray}{(STD)}  & LPIPS  $\downarrow$ \textcolor{gray}{(STD)}   \\
        \midrule
      Realfusion~\cite{melas2023realfusion}   
        & 15.26 \textcolor{gray}{\textsuperscript{$\pm$ 0.24}} & 0.722 \textcolor{gray}{\textsuperscript{$\pm$ 0.017}} & 0.283 \textcolor{gray}{\textsuperscript{$\pm$ 0.005}}   \\
        Zero123~\cite{liu2023zero}   
        & 18.93 \textcolor{gray}{\textsuperscript{$\pm$ 0.22}} & 0.779 \textcolor{gray}{\textsuperscript{$\pm$ 0.015}} & 0.166 \textcolor{gray}{\textsuperscript{$\pm$ 0.005}}   \\
        SyncDreamer~\cite{liu2023syncdreamer}
        & {20.05 \textcolor{gray}{\textsuperscript{$\pm$ 0.22}}} & {0.798 \textcolor{gray}{\textsuperscript{$\pm$ 0.013}}} & {0.146 \textcolor{gray}{\textsuperscript{$\pm$ 0.004}}} \\
        SV3D~\cite{voleti2024sv3d}
        & {21.26 \textcolor{gray}{\textsuperscript{$\pm$ 0.26}}} & {0.880 \textcolor{gray}{\textsuperscript{$\pm$ 0.015}}} & {0.080 \textcolor{gray}{\textsuperscript{$\pm$ 0.005}}} \\
        Unique3D~\cite{wu2024unique3d}
        & {21.71 \textcolor{gray}{\textsuperscript{$\pm$ 0.19}}} & {0.913 \textcolor{gray}{\textsuperscript{$\pm$ 0.012}}} & {0.083 \textcolor{gray}{\textsuperscript{$\pm$ 0.003}}} \\
        Era3D~\cite{li2024era3d}
        & {22.49 \textcolor{gray}{\textsuperscript{$\pm$ 0.24}}} & {0.916 \textcolor{gray}{\textsuperscript{$\pm$ 0.011}}} & {0.069 \textcolor{gray}{\textsuperscript{$\pm$ 0.004}}} \\
        Wonder3D~\cite{long2024wonder3d}
        & 22.92 \textcolor{gray}{\textsuperscript{$\pm$ 0.22}} &  0.919 \textcolor{gray}{\textsuperscript{$\pm$ 0.009}} & 0.063 \textcolor{gray}{\textsuperscript{$\pm$ 0.002}} \\
        Ours-(Perspective Setting)    
        & 23.23 \textcolor{gray}{\textsuperscript{$\pm$ 0.21}} & 0.929 \textcolor{gray}{\textsuperscript{$\pm$ 0.012}} & 0.054 \textcolor{gray}{\textsuperscript{$\pm$ 0.003}} \\
        Ours-(Orthogonal Setting)    
        & \textbf{23.95 \textcolor{gray}{\textsuperscript{$\pm$ 0.26}}} & \textbf{0.933  \textcolor{gray}{\textsuperscript{$\pm$ 0.010}}} & \textbf{0.051 \textcolor{gray}{\textsuperscript{$\pm$ 0.003}}} \\
        \bottomrule
    \end{tabular}
    }
    \label{tab:nvs}
\vspace{-4mm}
\end{table}

%% file: sections/5_exp.tex
\section{Experiments}

\subsection{Implementation Details}
We train our model on a high-quality single object subset of the Objaverse dataset~\cite{deitke2023objaverse}, which comprises approximately 100k objects following a cleanup process. 
To support the model in learning consistent representations across diverse camera settings, we create the rendered augmented multi-view dataset. We first normalized each object to be centered and on a unit scale. Then we render normal maps and color images from six views, including the front, back, left, right, front-right, and front-left views, using Blenderproc~\cite{Denninger2023} under both orthogonal and perspective camera settings.

We fine-tune our multi-view cross-domain model starting from the Stable Diffusion Image Variations Model, which has previously been fine-tuned with image conditions. 
During fine-tuning, we use a reduced image size of 256 $\times$ 256 and a total batch size of 512 for training. The fine-tuning process involves training the model for 80,000 steps. This entire training procedure typically requires approximately 6 days on a cluster of 8 Huawei Kunpeng 910B GPUs.

For training the cross-domain enhancement module, we use rendered multi-view images at 512$\times$512 resolution as targets, enabling the model to learn robust multi-view cross-domain enhancement capabilities. The module was trained for 80,000 steps with a batch size of 24 on 8 Huawei Kunpeng 910B GPUs, completing in approximately two days.

\input{figures/ablation-cascaded-recon}

\subsection{Baselines}
We adopt a comprehensive set of baseline methods across three key categories: SDS-based methods, multi-view-based methods, and feed-forward methods. 
For SDS-based methods, RealFusion~\cite{melas2023realfusion} and Magic123~\cite{qian2023magic123} leverage Score Distillation Sampling (SDS) loss for guiding single-view 3D reconstruction from an input image.
In the MV-based category, Zero123~\cite{liu2023zero} generates novel views from arbitrary perspectives, which can be further combined with SDS loss~\cite{poole2022dreamfusion} to perform 3D reconstruction. Additionally, SyncDreamer~\cite{liu2023syncdreamer}, Era3D~\cite{li2024era3d}, and Wonder3D~\cite{long2024wonder3d} produce multi-view consistent images, using the Neus~\cite{wang2021neus} representation to reconstruct 3D geometry. In contrast, Unique3D~\cite{wu2024unique3d} directly optimizes mesh vertices for a reconstruction approach.
For feed-forward methods, Point-E~\cite{nichol2022point} and Shap-E~\cite{jun2023shap} are 3D generative models trained on a large internal OpenAI 3D dataset, both of which are able to convert a single-view image into a point cloud or an implicit representation. One-2-3-45~\cite{liu2023one}, CRM~\cite{wang2024crm} and InstantMesh~\cite{xu2024instantmesh} first generate multi-view results and then leverage a sparse-view feed-forward reconstruction model to deliver a textured mesh using the representation of triplane or implicit SDF representation.

\subsection{Evaluation Protocol}
\noindent
\textbf{Evaluation Datasets.} Following prior research~\cite{liu2023zero,liu2023syncdreamer}, we adopt the Google Scanned Object dataset~\cite{downs2022google} for our evaluation, which includes a wide variety of common everyday objects. 
Our evaluation dataset aligns with that of SyncDreamer~\cite{liu2023syncdreamer}, comprising 30 objects spanning from everyday items to animals.
For each object in the evaluation set, we render an image with a size of 256×256, which serves as the input. Additionally, we include in-the-wild images with diverse styles, collected from the internet or generated using text-to-image models, in our evaluation.

\noindent
\textbf{Metrics.} To evaluate the quality of the single-view reconstructions, we adopt two metrics: Chamfer Distances (CD) and Volume IoU between ground-truth shapes and reconstructed shapes on both orthogonal and perspective settings of our method.
Since different methods adopt various canonical systems, we first align the generated shapes to the ground-truth shapes before calculating the two metrics.
Moreover, we adopt the metrics PSNR, SSIM~\cite{wang2004image} and LPIPS~\cite{zhang2018unreasonable} for evaluating the generated color images. 

\subsection{Single View Reconstruction}

We compare the quality of the reconstructions with SOTA methods, the quantitative results are summarized in Table~\ref{tab:recon}, and the qualitative comparisons are presented in Fig.~\ref{gso_compare}. 
Rather than modeling multi-view images and normal maps distribution jointly, Unique3D~\cite{wu2024unique3d} sequentially generate multi-view colors and normal maps. This exacerbates the geometric inconsistency since the stage 2 model has a huge domain gap in the inputs during training and in the inference period(see Section \ref{sec:discussion} for a more detailed discussion).
As a result, vertices are updated by unstable and incorrect gradients during the optimization process, destabilizing the process and ultimately leading to a reconstructed object with incorrect shapes.
Leveraging the generalization performance of the multi-view diffusion model and sparse-view reconstruction model, InstantMesh~\cite{xu2024instantmesh} produces results with reasonable overall geometry. However, due to the limited resolution of the triplane representation, the generated geometry and textures lack sufficient detail and exhibit artifacts aligned perpendicular to the coordinate axes.
Wonder3D~\cite{long2024wonder3d} takes relatively low-resolution multi-view images and normal maps to reconstruct 3D objects using implicit SDF representation without an enhancement process, it also struggles to obtain fine geometric and texture details.

Thanks to our multi-stage training scheme and cascaded 3D mesh extraction, our method produces textured meshes with a well-defined overall shape and superior geometric and textural quality compared to baseline approaches. Furthermore, geometric quantization comparisons on the GSO dataset~\cite{downs2022google} highlight the generalization ability and effectiveness of our method (see Table ~\ref{tab:recon}).

\subsection{Novel View Synthesis}
\input{figures/geo_refine}
We evaluate the quality of novel view synthesis for different multi-view-based methods. The quantitative results are presented in Table~\ref{tab:nvs}, and the qualitative results can be found in Figure~\ref{color-compare}. 
SyncDreamer~\cite{liu2023zero} introduces a volume attention scheme to enhance the consistency of multi-view images, their model is sensitive to the elevation degrees of the input images and tends to produce unreasonable results.
Unique3D~\cite{wu2024unique3d} stitching together the multiple views into a single image to generate multi-view results. This method severely limits the number of viewpoints it can produce (up to four). Moreover, generating images and normal maps in a sequential manner limits the generalizability and cross-view consistency of the model.
Era3D~\cite{li2024era3d} proposes row-wise attention to ensure multi-view consistency and estimates the input camera focal length to normalize the output results into an orthogonal coordinate system. Although it generates results with a higher resolution, it is difficult to stably output reasonable multi-view results for in-the-wild images.

In contrast, by learning information from both the color and normal domains simultaneously and exchanging information across domains via our cross-domain attention layers, our cross-domain multi-view diffusion model effectively learns the joint distribution of multi-view images and normal maps through our proposed multi-stage training scheme. This allows it to stably generate high-fidelity, multi-view consistent results.

\subsection{Discussions}
\label{sec:discussion}
\input{figures/cam_type_vis}

In this section, we perform a series of experiments to validate the effectiveness of the designs in our method. 

\noindent
\textbf{Cross-Domain Diffusion.}
To validate the effectiveness of our proposed cross-domain diffusion scheme, we study the following settings: (a) cross-domain model with cross-domain attention; 
(b) cross-domain model without cross-domain attention; 
(c) sequential models normal-to-rgb: first train a multi-view color diffusion model then train a multi-view normal diffusion model conditioned on the previously generated color images; (d) sequential models rgb-to-normal: first train a multi-view normal diffusion model then train a multi-view color diffusion model conditioned on the previously generated normal images.

As shown in (a) and (b) of Figure~\ref{fig:ablation_condx}, it's evident that the cross-domain attentions significantly enhance the consistency between color images and normals, particularly in terms of the detailed geometries of objects like the ice-cream and the sculpture.
For the sequential model normal-to-rgb, conditioning on the separately generated normal maps, the generated color images exhibit color aberrations in comparison to the input image, as shown in (c) of Figure~\ref{fig:ablation_condx}. 
Conversely, for the sequential model rgb-to-normal, conditioning on the separately generated color images, the normal maps give unreasonable geometry, as illustrated in (d) of Figure~\ref{fig:ablation_condx}.
The result further substantiates our hypothesis in Sec.~\ref{sec:cross-domain}, confirming that the domain gap in stage 2 does contribute to performance degradation. These experiments also indicate that jointly predicting normal maps and color images via the cross-domain attention mechanism enables a more integrated perception of information across domains, effectively enhancing the model’s ability to capture consistent details and address domain-specific discrepancies.

\noindent
\textbf{Cascaded 3D Mesh Reconstruction.}
As visualized in Figure~\ref{fig:cascaded}, we present the results of each stage of our proposed cascaded pipeline to demonstrate the effectiveness.
(1) In the geometric initialization stage, we check whether the target mesh has a concave overall geometry to decide between using a sphere mesh or a Poisson-reconstructed mesh as the starting point for optimization. This ensures that the initial mesh attains an approximate but correct geometric topology, providing a robust foundation for efficient optimization and enabling our method to handle meshes with complex topologies (see (b) in Figure~\ref{fig:cascaded}).
(2) The inconsistency-aware coarse reconstruction method derives coarse textured meshes that capture preliminary geometric details and model multi-view inaccuracies and blurry explicit on the mesh (see Figure~\ref{fig:cascaded}).
(3) Finally, within the iterative refinement stage, enhanced multi-view outputs are employed to further refine the mesh’s geometry and texture. As a result, the mesh surface gains additional detail, effectively correcting geometric errors introduced by multi-view inaccuracy (see (d) in Figure~\ref{fig:cascaded}).

\noindent
\textbf{Discussion with Wonder3D.}
We also include a comparison about the reconstruction strategy with wonder3D.
As shown in Figure~\ref{fig:recon_method}, our approach produces smoother, more detailed meshes with enhanced geometric and texture fidelity. This improvement is attributed to: 1) the cascading structure, which supports a coarse-to-fine 3D object extraction, avoid detail loss that often occurs during transformations between mesh and SDF representation; and 2) our cross-domain multi-view enhancement module, which iteratively refines both geometry and texture, achieving higher resolution while correcting viewpoint inconsistencies, resulting in superior overall quality.

\input{figures/recon_method}

\noindent
\textbf{Geometry-aware Normal Loss.}
To validate the effectiveness of our proposed geometry-aware normal loss, we conduct an experiment to reconstruct geometry with or without such loss.
As shown in Figure~\ref{fig:geo_refine}, the reconstructed geometry with geometry-aware loss presents more geometric details and avoids surface irregularities caused by inconsistent gradients. This is because the proposed loss could further alleviate the inaccuracies among the generated multi-view normal maps, enhancing the geometric detail of the mesh surface.

\noindent
\textbf{Camera Type Switcher.}
We perform experiments to validate the effectiveness of the camera type switcher, which allows our model to process input images captured by both perspective and orthogonal cameras. 
As shown in Figure~\ref{fig:cam_type_vis}, when input images from either orthogonal or perspective cameras are processed with an incorrectly configured camera type switcher, the reconstruction results exhibit significant distortions and degraded texture quality.
These experiments confirm that our proposed camera type switcher enables the model to achieve robust 3D generation across different imaging systems.

\input{figures/ablation-multi-stage-pt}
\noindent
\textbf{Multi-stage Training Strategy.} 
To discuss the efficiency of our multi-stage training strategy, we conduct an experiment that omits the multi-domain pre-training phase and directly fine-tunes the diffusion model on the cross-domain multi-view generation task. 
The qualitative results in Figure~\ref{fig:multi_stage} clearly demonstrate that incorporating multi-domain pre-training significantly improves both the generalization of structural consistency and the fidelity of generated multi-view outputs. 
Notably, we incorporate multi-view masks as part of the multi-domain pre-training, where input and output views are randomly replaced with masks. Although predicting multi-view masks is not a direct objective in the cross-domain generation task, these masks provide essential shape information about the 3D objects. This also confirms that our multi-stage training strategy effectively bridges the pre-trained image generation task and the cross-domain multi-view generation task in a step-by-step manner.

%% file: figures/ablation-cascaded-recon.tex
\begin{figure}[t]
\centering
\includegraphics[width=\linewidth]{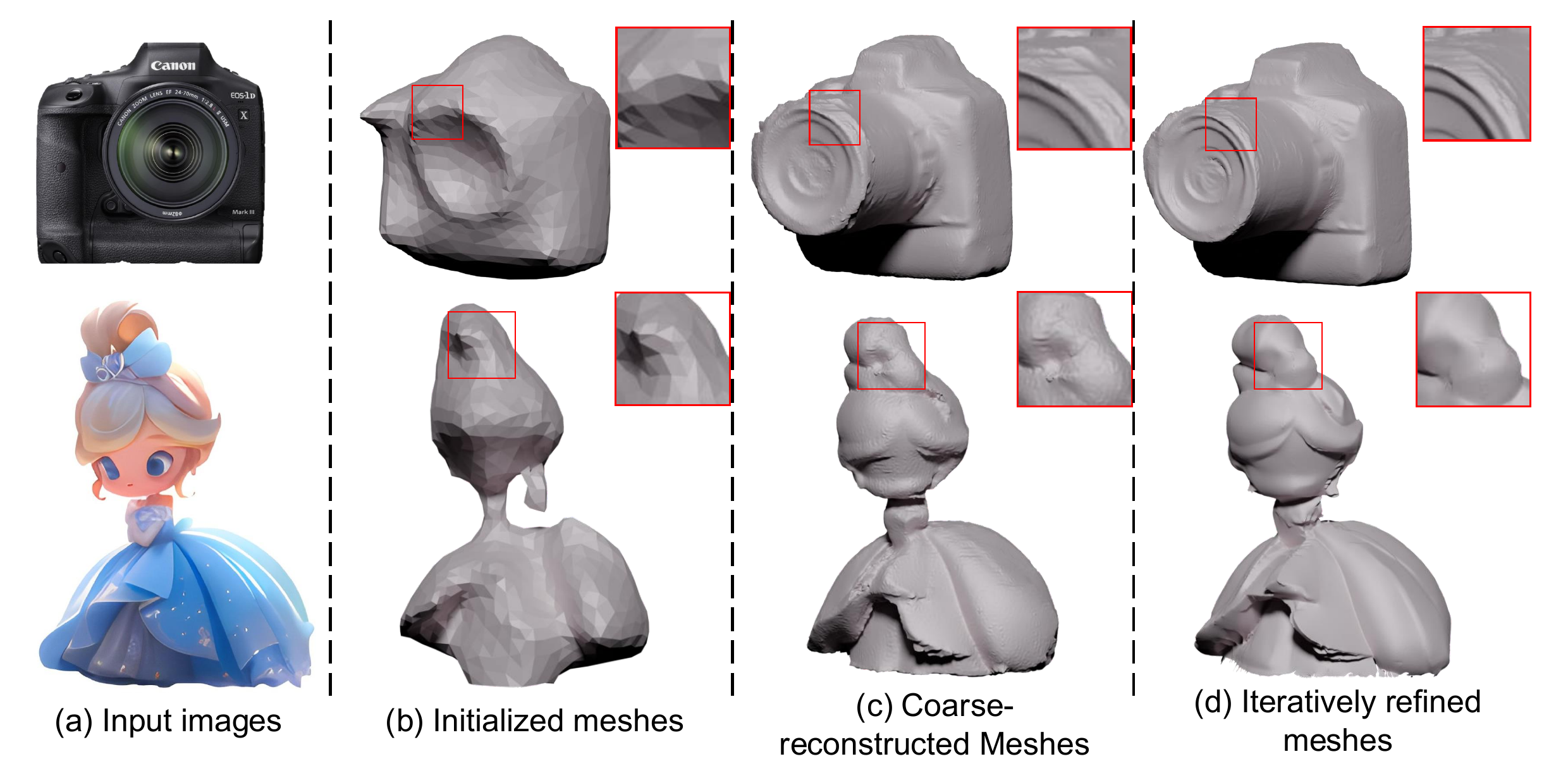}

\caption{Visualization of our cascaded 3D mesh extraction method shows how the geometry, from general topology to fine surface details, is captured in a coarse-to-fine manner, with subtle geometric errors corrected iteratively.
}
\vspace{-4mm}
\label{fig:cascaded}	
\end{figure}

%% file: figures/geo_refine.tex
\begin{figure}[tp!]
\centering
\includegraphics[width=1\linewidth]{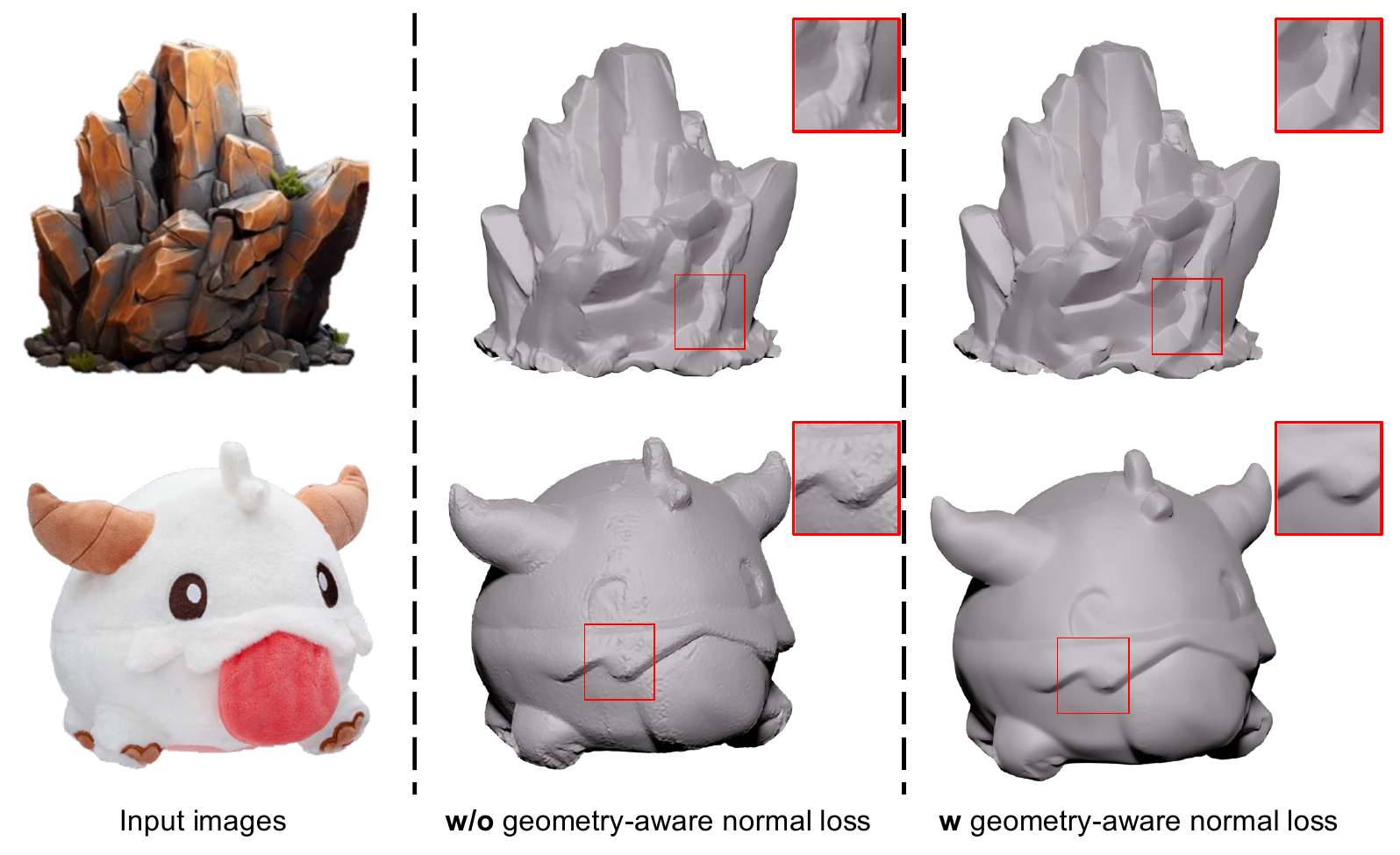}
\caption{Geometric visualization of the effect of our geometry-aware normal loss. Our geometry-aware normal loss effectively mitigates inaccuracies in the generated normal maps, enhancing the surface details of the meshes.}
\vspace{-5mm}
\label{fig:geo_refine}	
\end{figure}

%% file: figures/cam_type_vis.tex
\begin{figure}[tp!]
\centering
\includegraphics[width=\linewidth]{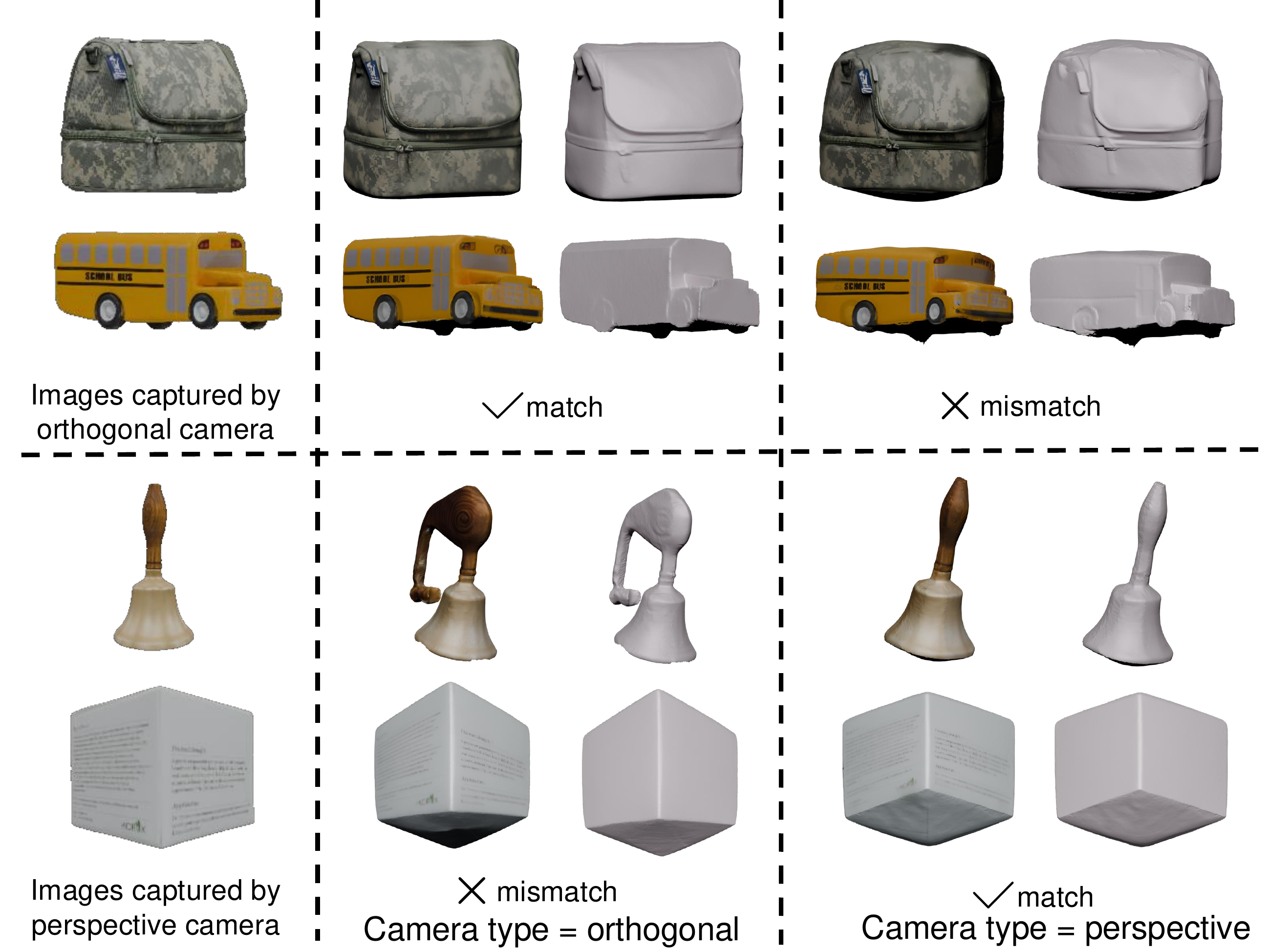}
\caption{Ablation study on our camera type switcher. The camera type switcher enables the generation process across various projection types, thereby enhancing the robustness of our method.}
\vspace{-6mm}
\label{fig:cam_type_vis}	
\end{figure}

%% file: figures/recon_method.tex
\begin{figure}[tp!]
\centering
\includegraphics[width=1\linewidth]{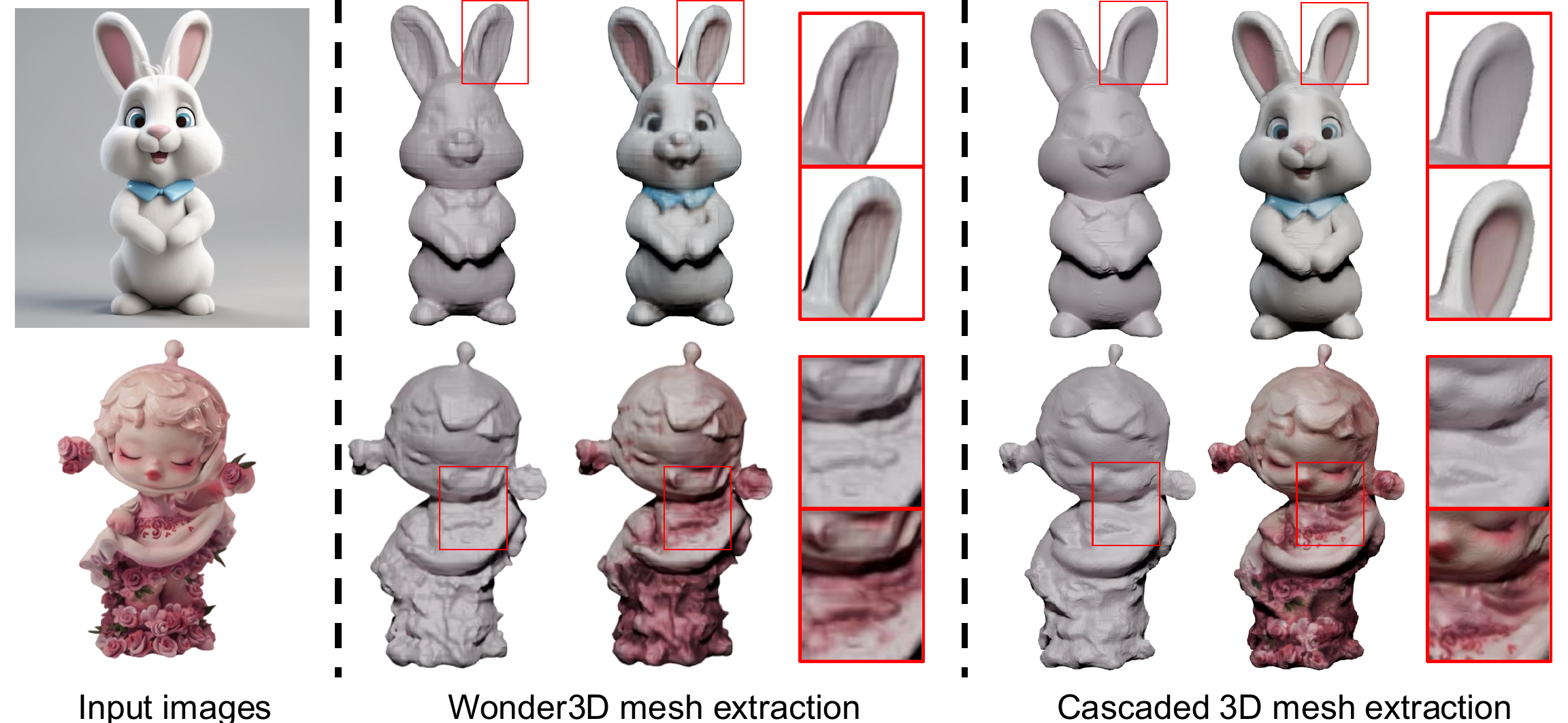}
\caption{The qualitative comparisons of reconstruction method between the geometry fusion algorithm proposed by Wonder3D~\cite{long2024wonder3d} and cascaded 3D mesh extraction method. Our approach has significant advantages in both geometric and textural details.}
\vspace{-6mm}

\label{fig:recon_method}	
\end{figure}

%% file: figures/ablation-multi-stage-pt.tex
\begin{figure}[tp!]
\centering
\includegraphics[width=\linewidth]{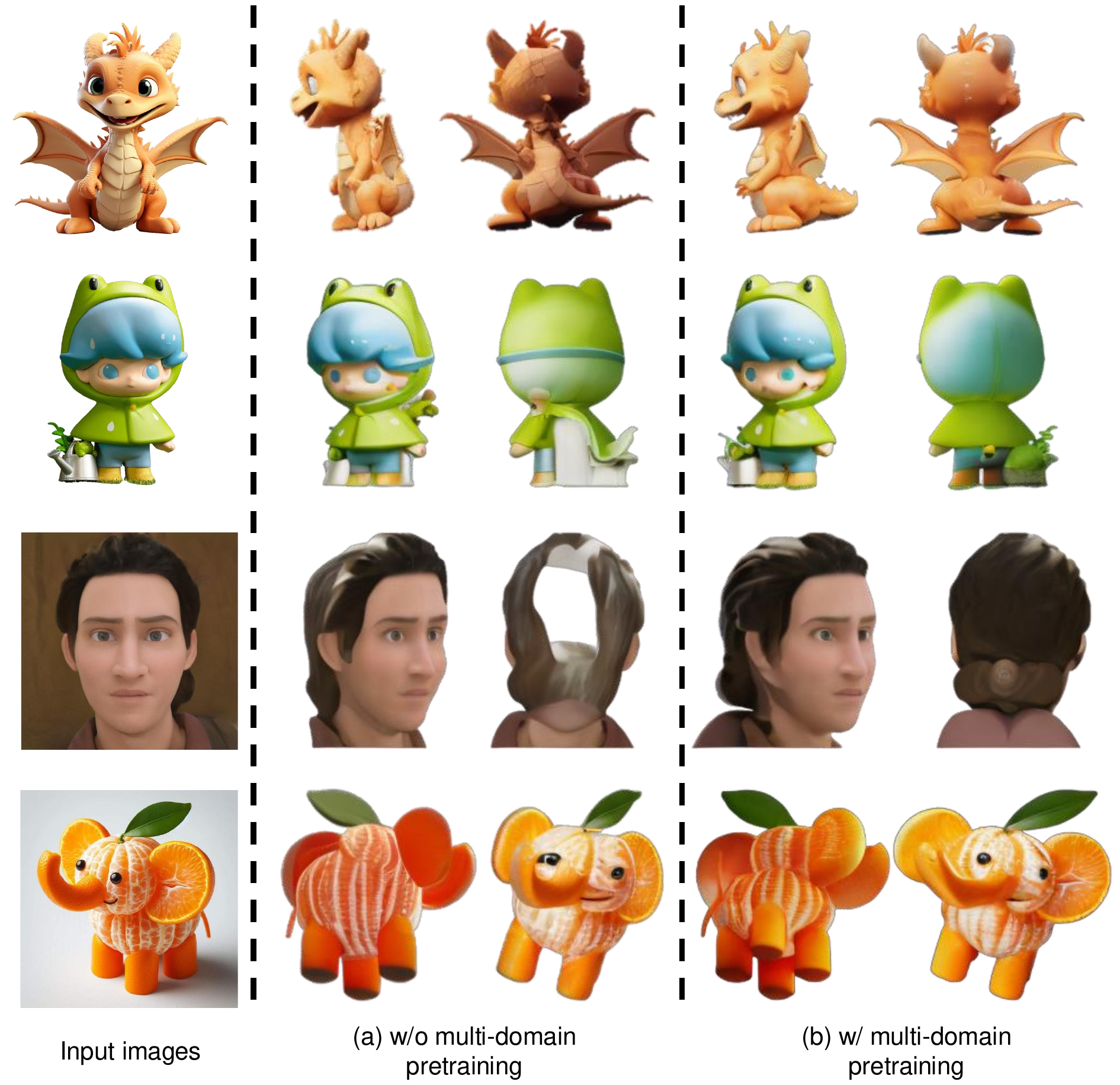}
\caption{Ablation study of the multi-stage training strategy. The results demonstrate that our multi-domain pretraining significantly improves both the consistency of the overall shape and the fidelity of the generated results.}
\vspace{-6mm}
\label{fig:multi_stage}	
\end{figure}

%% file: sections/6_conclusion.tex
\section{Conclusions and Limitations}

In this paper, we present \textit{Wonder3D++}, an innovative framework for efficiently generating high-fidelity textured meshes from single-view images. Given an input image, \textit{Wonder3D++} leverages a cross-domain multi-view diffusion module to generate multi-view images and normal maps. These results are then utilized in a novel cascaded 3D mesh extraction algorithm, consisting of geometric initialization, inconsistency-aware coarse reconstruction, and an iterative refinement stage to extract textured mesh. Experimental results demonstrate that our method achieves robust generalization and produces textured meshes with superior geometric and textural quality.

While \textit{Wonder3D++} has shown promising performance in reconstructing 3D geometry from single-view images, it encounters challenges when handling objects with highly complex geometries and severe self-occlusion. This limitation arises from the restricted number of viewpoints and the inherent constraints of the mesh-based optimization method. We believe these challenges can be alleviated by increasing the number of viewpoints and adopting a more robust mesh-based reconstruction strategy.

%% file: sections/acknowledgment.tex
\section*{Acknowledgment}
This work was supported by Project NSFC 62501261.
Additionally, this research was also supported by the Lhasa Science and TechnologyProgram Project titled
"Empowering Grassroots Social Governance Modernization through 5G and Multimodal Big Data Technologies"(Project No.LSKJ-202402).

%% file: sections/bio.tex
\section{Biography Section}
\begin{IEEEbiography}[{\includegraphics[width=1in,height=1.25in,clip,keepaspectratio]{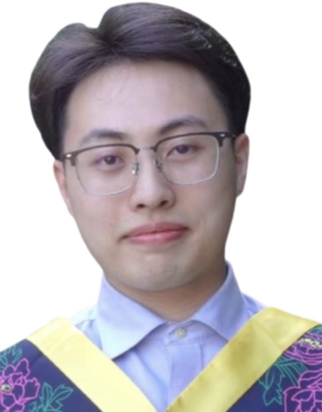}}]{Yuxiao Yang}
received his B.S. degree in Information Engineering from Zhejiang University, Hangzhou, China, in 2023. He is currently pursuing a master’s degree at the Shenzhen International Graduate School, Tsinghua University. His research interests include 3D generation and video generation.
\end{IEEEbiography}
\begin{IEEEbiography}[{\includegraphics[width=1in,height=1.25in,clip,keepaspectratio]{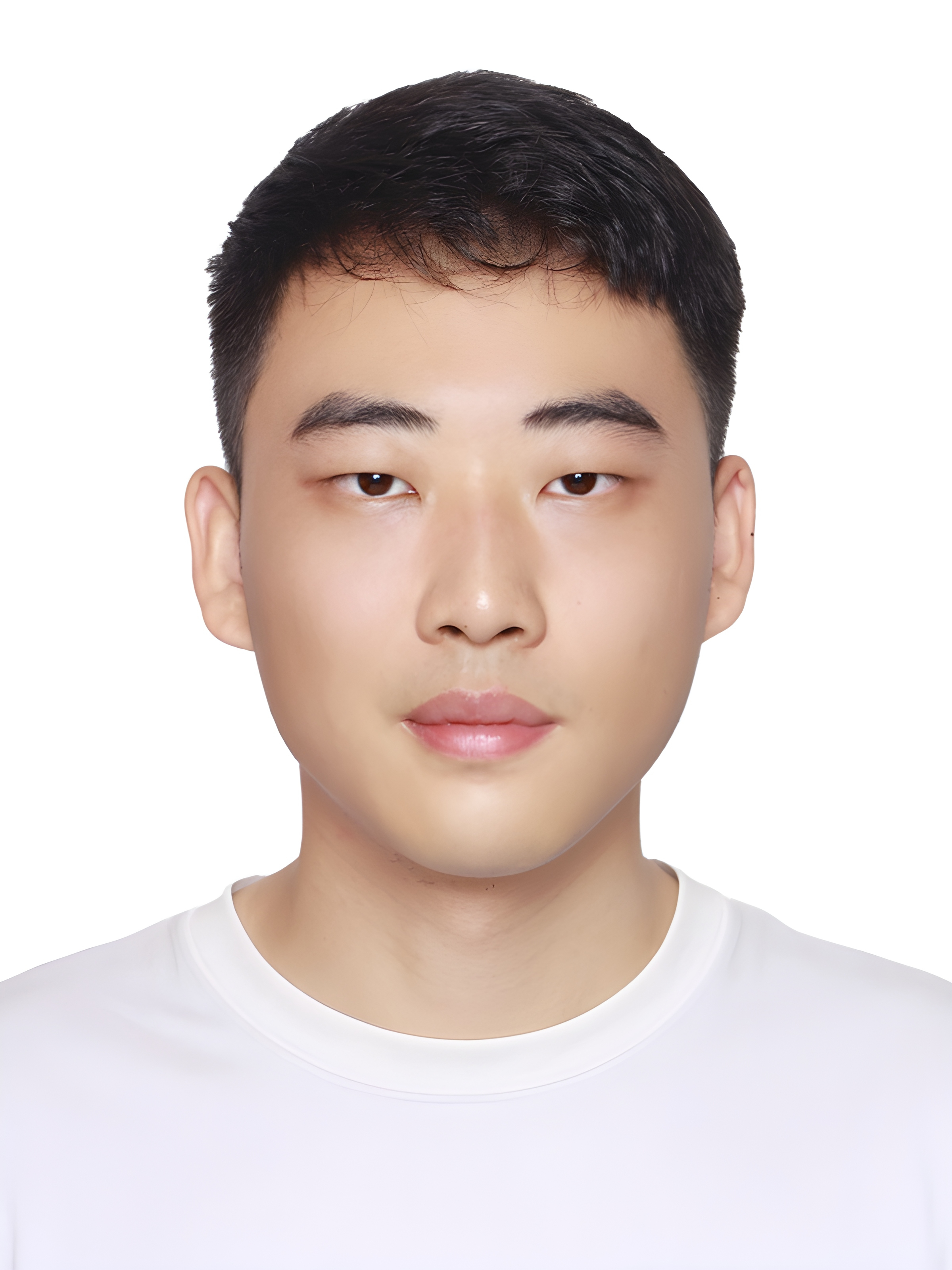}}]{Xiaoxiao Long} is an Tenure-Track Associate Professor at School of Intelligence Science and Technology, Nanjing University. He obtained his Ph.D degree at the University of Hong Kong in 2023. His research includes 3D vision, computer graphics and robotic AI.
\end{IEEEbiography}
\begin{IEEEbiography}[{\includegraphics[width=1in,height=1.25in,clip,keepaspectratio]{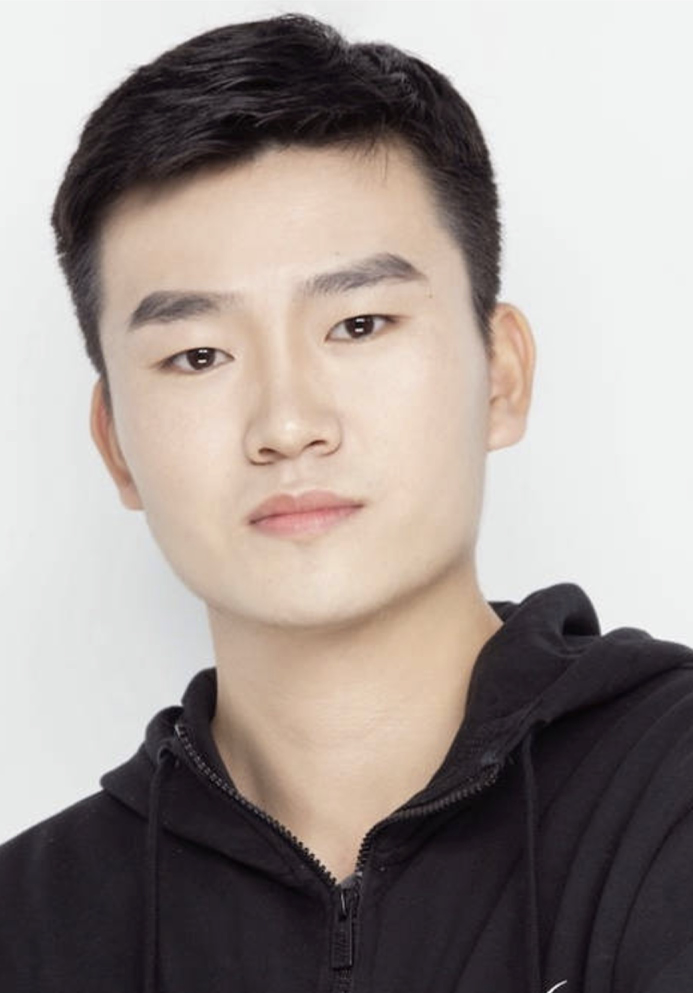}}]{Zhiyang Dou} is a Ph.D. candidate in the Computer Graphics Group at the University of Hong Kong, supervised by Prof. Wenping Wang and Prof. Taku Komura. He received his B. Eng. degree with honors from Shandong University, advised by Prof. Shiqing Xin. He is currently a visiting Ph.D. student in the Department of Computer and Information Science at the University of Pennsylvania, working at the Graphics Lab and GRASP Lab.
\end{IEEEbiography}
\begin{IEEEbiography}[{\includegraphics[width=1in,height=1.25in,clip,keepaspectratio]{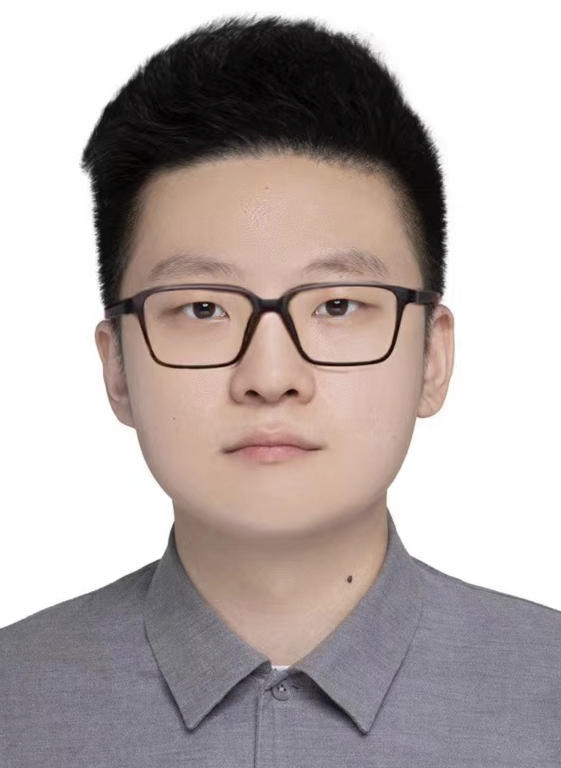}}]{Cheng Lin} received his Ph.D. from The University of Hong Kong (HKU), advised by Prof. Wenping Wang. Now he is an assistant professor at Macau University of Science and Technology. He visited the Visual Computing Group at Technical University of Munich (TUM), advised by Prof. Matthias Nießner. Before that, he completed his B.E. degree at Shandong University. His research interests include geometric modeling, 3D vision, shape analysis, and computer graphics.
\end{IEEEbiography}
\begin{IEEEbiography}[{\includegraphics[width=1in,height=1.25in,clip,keepaspectratio]{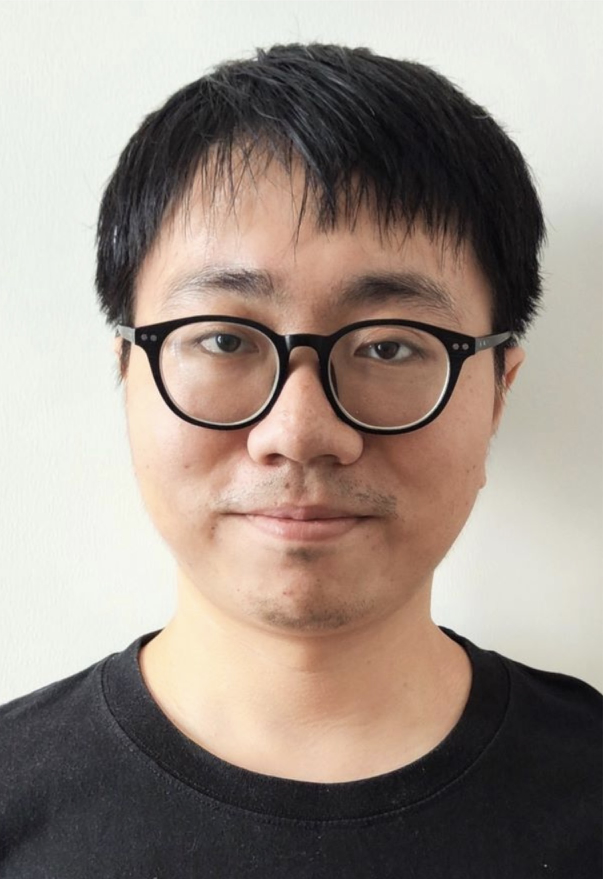}}]{Yuan Liu} is visiting NTU as a PostDoc working with Prof. Ziwei Liu. He obtained his PhD degree at HKU advised by Prof. Wenping Wang and Prof. Taku Komura. His research mainly concentrates on 3D vision and graphics. He currently work on topics about 3D AIGC including neural rendering, neural representations, and 3D generative models.
\end{IEEEbiography}
\begin{IEEEbiography}[{\includegraphics[width=1in,height=1.25in,clip,keepaspectratio]{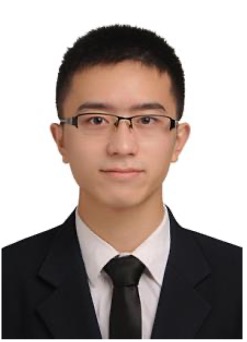}}]{Qingsong Yan} is currently a Ph.D. student at the School of Geodesy and Geomatics, Wuhan University. He received the B.Eng. degree and MSc degree from Wuhan University in 2016 and 2019, respectively. He is majoring in omnidirectional computer vision, 3D reconstruction, multiview-stereo, and neural implicit representations.
\end{IEEEbiography}
\begin{IEEEbiography}[{\includegraphics[width=1in,height=1.25in, clip,keepaspectratio]{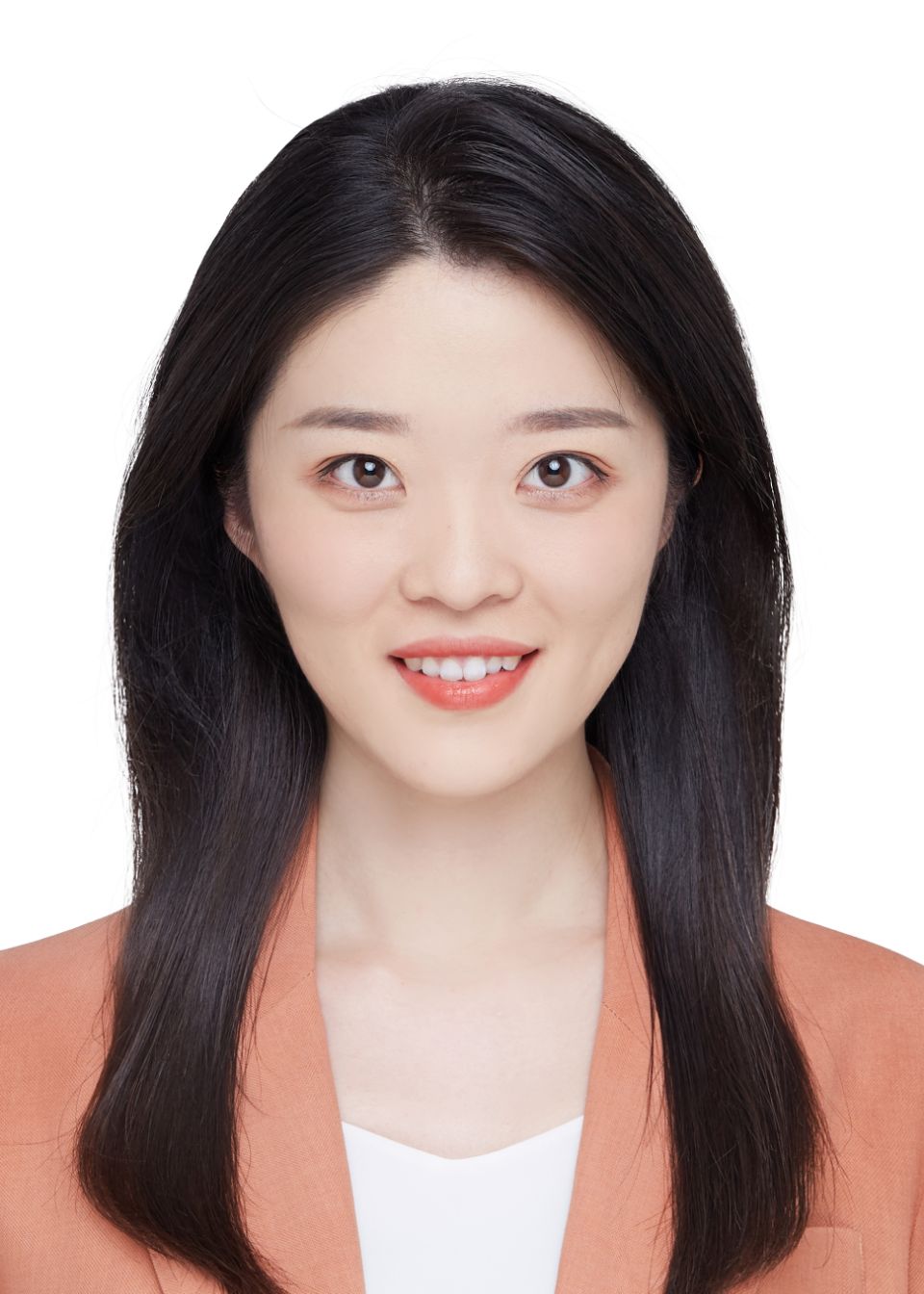}}]{Yuexin Ma} received the PhD degree from the Department of Computer Science at University of Hong Kong in 2019.
Now she is an assistant professor in ShanghaiTech University, China, leading 4DV Lab. Her research interests lie on the computer vision and deep learning. Particularly she is interested in 3D scene understanding, multi-modal perception, human-machine cooperation, and autonomous driving. She has published dozens of top-tier journal and conference papers in computer vision and robotics, including Science Robotics, TPAMI, IJCV, CVPR, ECCV, AAAI, TOG, TVCG, SIGGRAPH, ICRA, etc.
\end{IEEEbiography}
\begin{IEEEbiography}[{\includegraphics[width=1in,height=1.25in,clip,keepaspectratio]{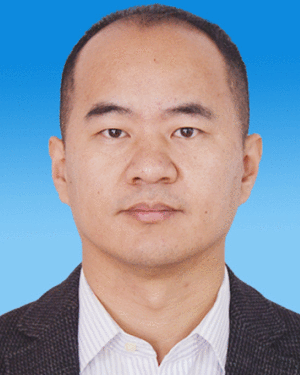}}]{Haoqian Wang}
(Member, IEEE) received the PhD degree from the Harbin Institute of Technology,
Harbin, in 2005. He was a post-doctoral fellow with Tsinghua University, Beijing, China, from 2005 to 2007. He has been a faculty member with the Shenzhen International Graduate School, Tsinghua University, Shenzhen, China, since 2008, where he has also been an associate professor and a professor since 2011 and the director of the Shenzhen Institute of Future Media Technology. His current research interests include Stereo Vision and deep learning.
\end{IEEEbiography}
\begin{IEEEbiography}[{\includegraphics[width=1in,height=1.25in,clip,keepaspectratio]{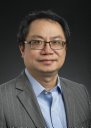}}]{Zhiqiang Wu}
is the Brage Golding Distinguished Professor of Electrical Engineering of Wright State University. He received his B.S. from Beijing University of Posts and Telecommunications in 1993, his M.S. from Peking University in 1996, and his Ph.D. from Colorado State University in 2002. He teaches courses in communication, signal processing and electronic warfare. His research interests include artificial intelligence, communication, signal processing, electronic warfare, and big data. His research has been funded by NSF, AFRL, AFOSR, ONR, NASA, DOE, and OFRN. He has published more than 200 papers in journals and conferences.  
\end{IEEEbiography}
\begin{IEEEbiography}[{\includegraphics[width=1in,height=1.25in,clip,keepaspectratio]{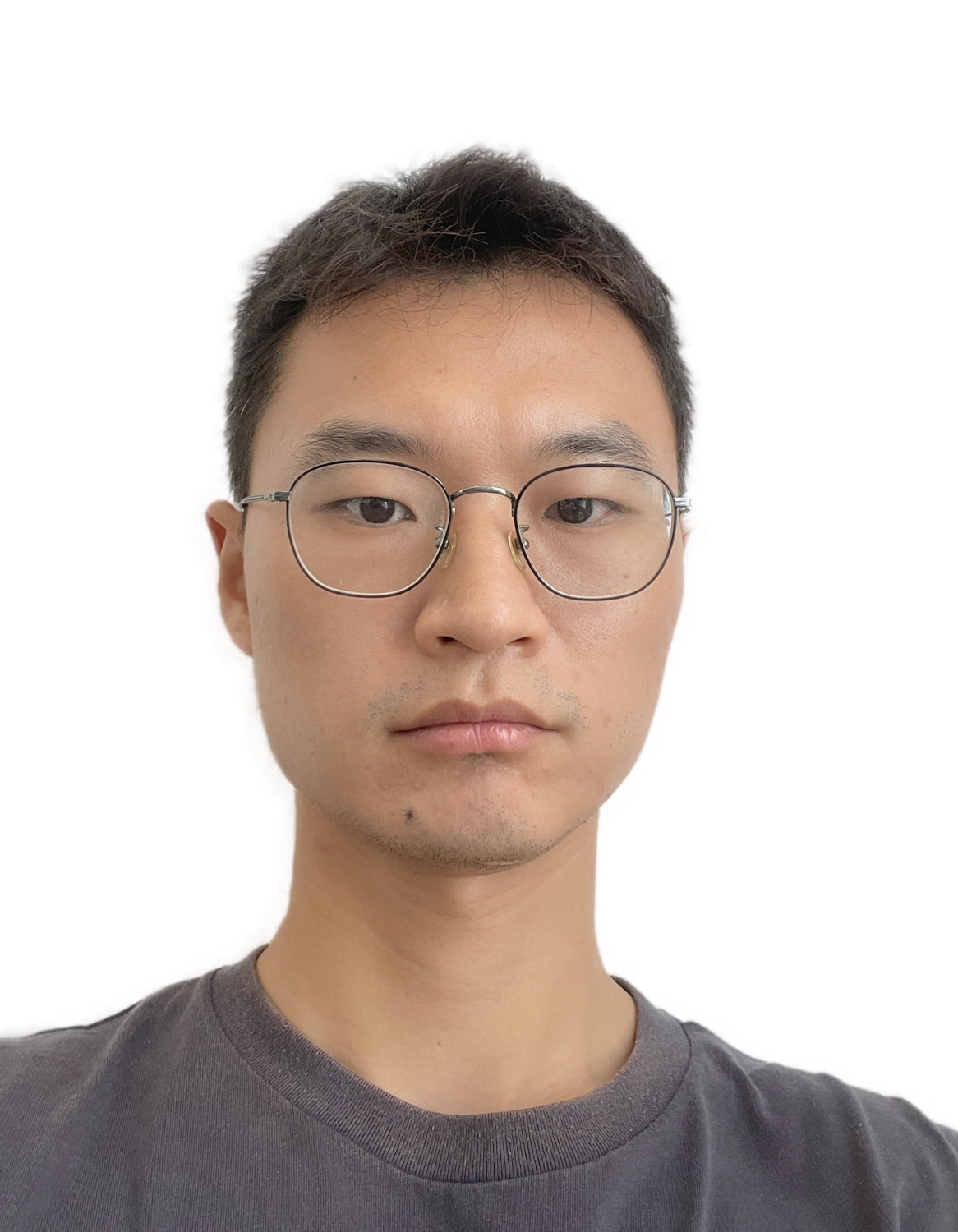}}]{Wei Yin}
is a staff research scientist at Horizon Robotics. Before that, he received Ph.D. degree at Australian Institure of Machine Learning (AIML), University of Adelaide. His research interests include the world model, 3D reconstruction, and depth estimation.
\end{IEEEbiography}
\vfill

%% file: sections/7_supp.tex
\section{Coordinate System}
\begin{figure*}[t!]
\centering
\includegraphics[width=0.95\textwidth]{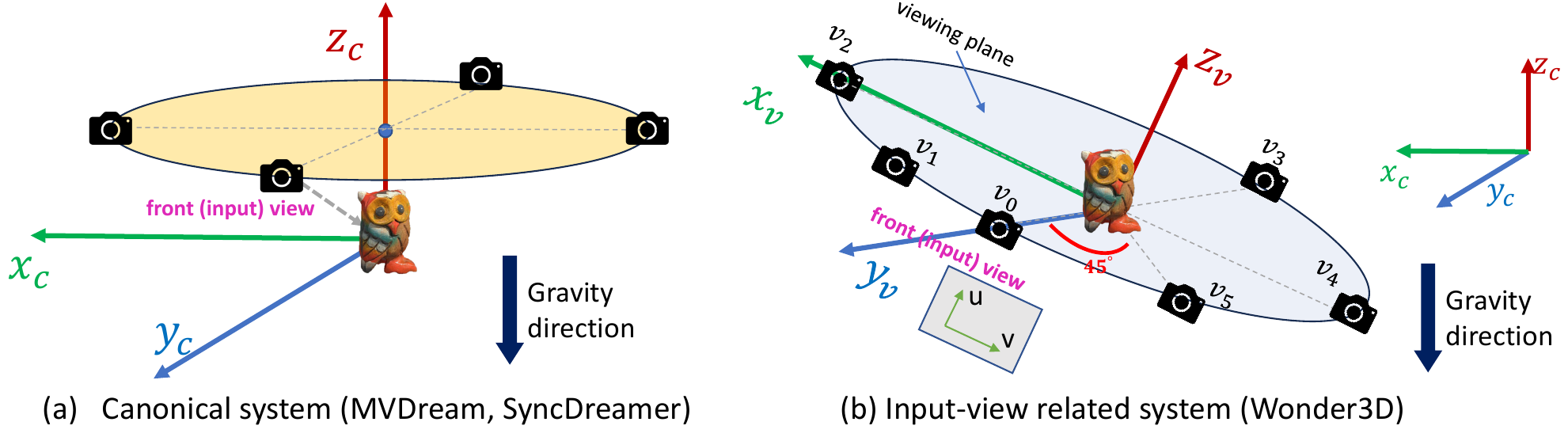}
\caption{Illustration of the coordinate systems and camera poses.}
\label{fig:coordinatee}	
\end{figure*}

In practice, the target object is assumed to be placed along the gravity direction.
\textbf{1) \textit{Canonical coordinate system.}} Some prior works (e.g. MVDream~\cite{shi2023mvdream} and SyncDreamer~\cite{liu2023syncdreamer}) adopt a shared canonical system for all objects, whose axis $Z_c$ shares the same direction with gravity (Fig.~\ref{fig:coordinatee} (a)). 
\textbf{2) \textit{Input view related system.}} Wonder3D and Wonder3D++ adopt an independent coordinate system for each object that is related to the input view.
Its $Z_v$ and $X_v$ axes are aligned with the UV dimension of 2D input image space, and its $Y_v$ axis is vertical to the 2D image plane and passes through the center of ROI (Region of Interests) (Fig.~\ref{fig:coordinatee} (b)).
\textbf{3) \textit{Camera poses.}} Wonder3D and Wonder3D++ outputs 6 views $\{v_i, i=0,...,5\}$ that are sampled at the $X_vOY_v$ plane of the input-view related system with a fixed radius, where the front view $v_0$ is initialized as input view and the other views are sampled with pre-defined azimuth degrees (see Fig.~\ref{fig:coordinatee} (b)).

\section{Effectiveness of cross-domain multi-view enhancement module}

To demonstrate the Effectiveness of our cross-domain multi-view enhancement module, in figure~\ref{fig:supp_enhance}, we visualize the multi-view results rendered by the coarse reconstructed mesh and the enhanced multi-view results. These results showcase that our cross-domain multi-view enhancement module effectively enhances texture and geometric blurring on the surface of the mesh.

\input{figures/supp_enhance}

\begin{figure}[t!]
\centering
\includegraphics[width=\columnwidth]{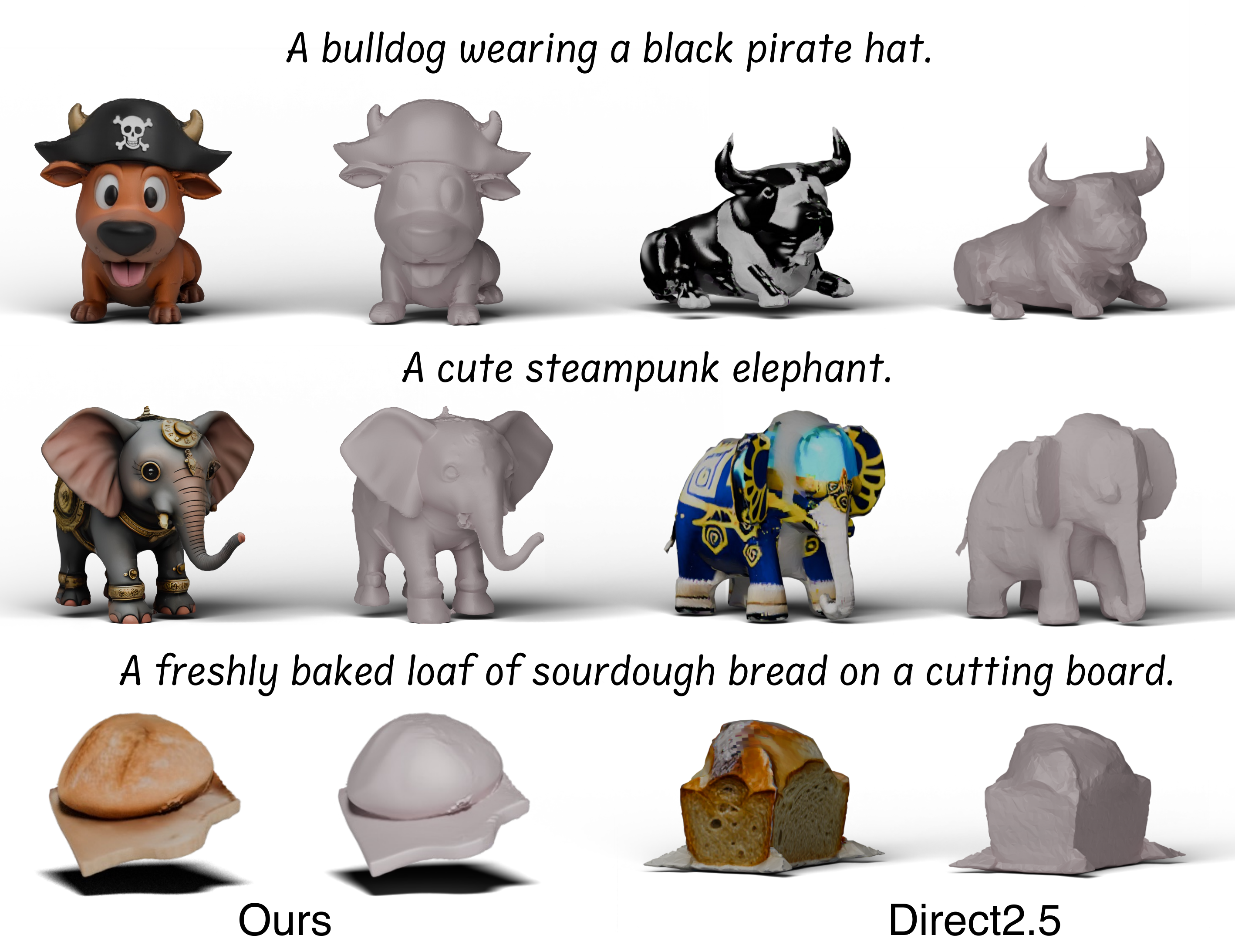}
\caption{Quantitative comparison with Direct2.5~\cite{lu2024direct2}}
\label{fig:t23d}	
\end{figure} 
\vspace{-4mm}

\section{More compassion on single image 3D mesh generation}

\begin{figure*}[t!]
\centering
\includegraphics[width=0.95\textwidth]{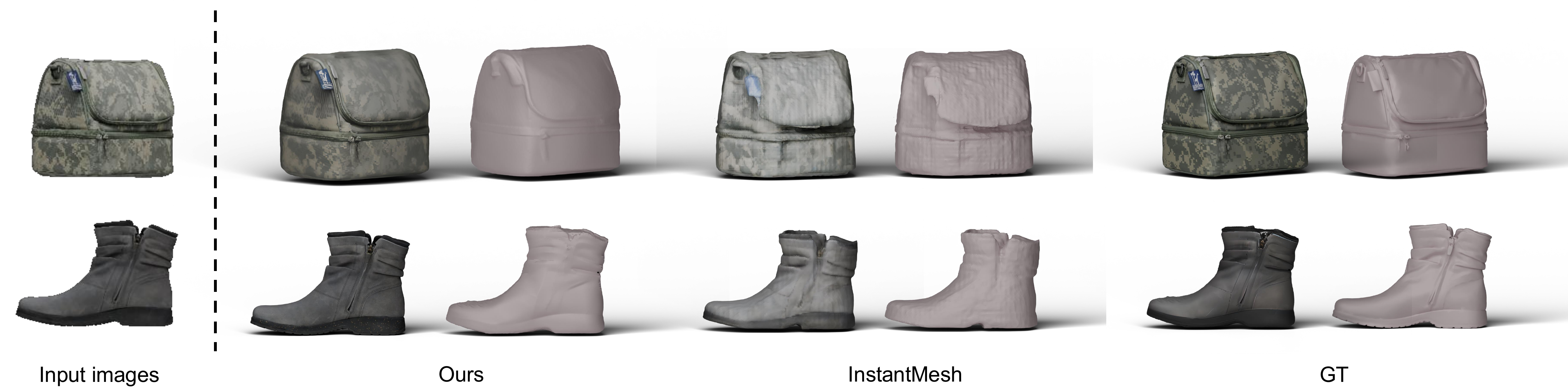}
\caption{Visual comparison of textured meshes generated by our method against ground truth.}
\label{fig:compare_gt}	
\end{figure*} 

\input{figures/supp_mesh_compare}

In this section, we showcase more results on single image 3D mesh generation task compared with InstantMesh~\cite{xu2024instantmesh} in figure~\ref{fig:supp_mesh_compare}. Our method generates meshes with better geometric details and texture fidelity.

\IEEEpubidadjcol

\section{Cross-domain novel view synthesis}

To demonstrate the effectiveness of our multi-stage training scheme on the cross-domain novel view synthesis task, we showcase the generated multi-view images and normal maps by our model and Wonder3D model in Fig ~\ref{fig:supp_nvs}.
Thanks to the addition of our multi-stage training method, our method significantly improves the realism of the RGB images, the rationalization of the shape of the object, and the accuracy of the normal maps.

\begin{figure*}[t!]
\centering
\includegraphics[width=\textwidth]{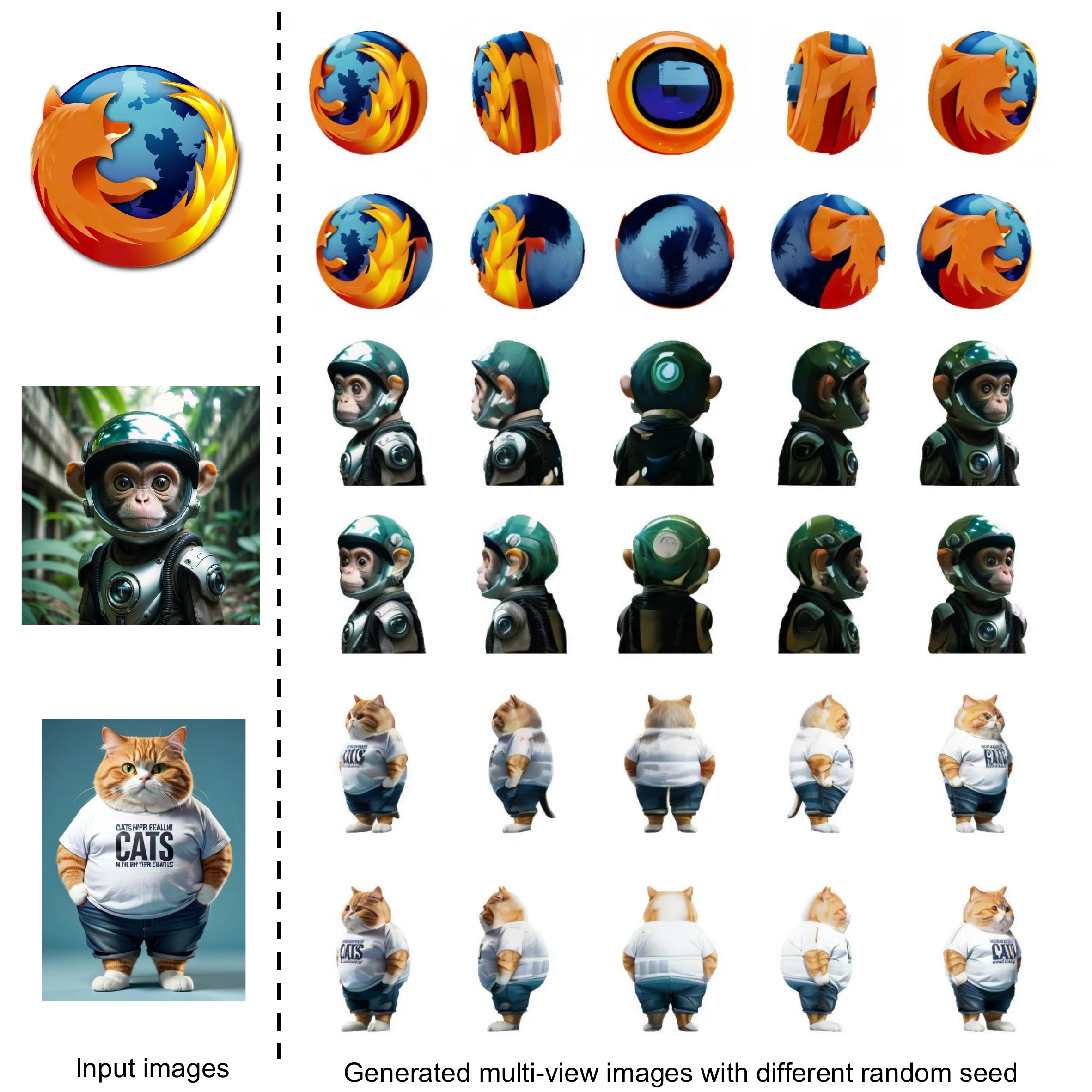}
\caption{Diversity of our generation process.}
\label{fig:diversity}	
\vspace{-4mm}
\end{figure*} 

\input{figures/supp_nvs}

\section{Incorporating information from the input image during the generation process.}
Our model is builded on a pre-trained Stable Diffusion image-to-image variations framework. Adopting it's design, we employ two parallel branches to inject visual guidance from the input image. 
 
Firstly, in order to guide the generative process using the input image and utilize the generative priors within the SD, we utilize the CLIP Vision Encoder to extract \textbf{high-level semantic features} from the input image, which are injected via Cross Attention layers.

Simultaneously, the second branch encodes the input image using a pre-trained VAE and concatenates the resulting latent features with the noise latent at the channel dimension, supplying \textbf{pixel-level visual information}.

\section{Inference Time}
To detail the computational cost of our method during the inference phase, Table~\ref{tab:timing} presents the time cost for each stage in the pipeline, measured on a single A100 GPU. Although our approach consists of multiple stages, it completes the generation process in a relatively short time, facilitated by the adaptation of remeshing optimization methods for modern GPUs.

\begin{table}[t]
\centering
\caption{Inference time of our pipeline.}
\label{tab:timing}
\begin{tabular}{lrrl}
\toprule
\textbf{Processing Stage} & \textbf{Time (s)} & \textbf{Percentage} \\
\midrule
Multi-View Generation & 28 & 17.39\% \\
Geometric Initialization & 3 & 1.86\%  \\
Coarse Reconstruction & 39 & 24.22\%  \\
Iterative Refinement & 91 & 56.52\%  \\
\midrule
\textbf{Total} & \textbf{161} & \textbf{100\%} & \\
\bottomrule
\end{tabular}
\end{table}

\section{Text-to-3D Generation}
While primarily designed for image-to-3D reconstruction, \textit{Wonder3D++}'s architecture allows for seamless integration with text-to-image diffusion models, such as Stable Diffusion~\cite{rombach2022high}, to enable text-driven 3D generation. This is achieved by first generating a single image conditioned on the input text and subsequently processing it through our pipeline to produce a 3D mesh. As demonstrated in Fig.~\ref{fig:t23d}, our method generates 3D assets exhibiting superior texture fidelity and geometric accuracy when compared to Direct2.5~\cite{lu2024direct2}. In contrast to our cross-domain joint sampling scheme, Direct2.5~\cite{lu2024direct2} employs a two-stage pipeline involving multi-view normal map generation followed by image generation for mesh extraction. This separation hinders the model from learning a coherent texture and geometry data distribution. A comprehensive numerical comparison is presented in Table \ref{tab:user_study_t23d_vs_direct2.5}.

\begin{table}[t]
\caption{User study results on generated 3D mesh quality. Abbreviations: w/o IR = without Iterative Refinement, w/o CTS = without Camera Type Switcher, w/o MDP = without Multi-domain Pretraining.}
\label{tab:user_study}
\centering
\begin{tabular}{lcccc}
\toprule
  & \textbf{Texture} & \textbf{Geometric} & \textbf{Shape} & \textbf{Overall} \\
 & \textbf{Quality} & \textbf{Details} & \textbf{Plausibility} & \textbf{Quality} \\
\midrule
Full Model & \textbf{8.25} & \textbf{8.81}  & \textbf{9.11}  & \textbf{8.79}  \\
w/o MDP & 7.35  & 6.23  & 7.14  & 8.15  \\
w/o CTS & 7.25  & 6.49  & 5.32  & 7.23  \\
w/o IR & 6.33  & 5.6  & 8.92  & 5.18 \\
\bottomrule
\end{tabular}
\vspace{-0.2cm}
\end{table}

\section{Comparison with Ground Truth.}
To evaluate the fidelity, detailed geometry, and texture quality of 3D meshes generated by our method, as well as its generalization capabilities, we conduct visual comparisons on examples from the GSO~\cite{downs2022google} dataset against both ground truth (GT) meshes and results from InstantMesh~\cite{xu2024instantmesh}. As depicted in Fig.~\ref{fig:compare_gt}, the meshes generated by our method maintain a high degree of similarity to the ground truth, while exhibiting only minor loss in fine details.

\begin{table}[t]
\caption{User study comparison with Direct2.5 on text-to-3D generation. We evaluate based on texture quality, geometric details, shape plausibility, and prompt following.}
\label{tab:user_study_t23d_vs_direct2.5}
\centering
\begin{tabular}{lcccc}
\toprule
Method & \textbf{Texture} & \textbf{Geometric} & \textbf{Shape} & \textbf{Prompt} \\
& \textbf{Quality} & \textbf{Details} & \textbf{Plausibility} & \textbf{Following} \\
\midrule
Ours & \textbf{7.81} & \textbf{8.66} & \textbf{8.92} & \textbf{9.32} \\
Direct2.5 & 4.18 & 5.89 & 6.21 & 8.11 \\
\bottomrule
\end{tabular}
\vspace{-0.2cm}
\end{table}

\section{Quantitative Ablation Results.}
    To validate the effectiveness of Iterative Refinement, Camera Type switching, and Multi-domain Pretraining, we conducted a user study with 15 participants evaluating 120 samples. Each participant assessed 30 randomly selected samples across four metrics: texture quality, geometric details, shape plausibility, and overall quality. As shown in Table \ref{tab:user_study}, the full model achieves the highest scores across all evaluation dimensions. Notably, the absence of multi-domain pretraining or camera type switching leads to implausible overall shapes, while removing iterative refinement substantially degrades both geometric and textural details.

\section{Diversity of Sampling.}
    We showcase the diversity of generated samples. In this experiment, we generate different samples from the same input image using different random seeds. The results, presented in Fig.~\ref{fig:diversity}, demonstrate that our method maintains the content diversity derived from the pre-trained diffusion model, while coherently matching the geometry and texture information of the given input image.

%% file: figures/supp_enhance.tex
\begin{figure}[]
\centering
\includegraphics[width=\linewidth]{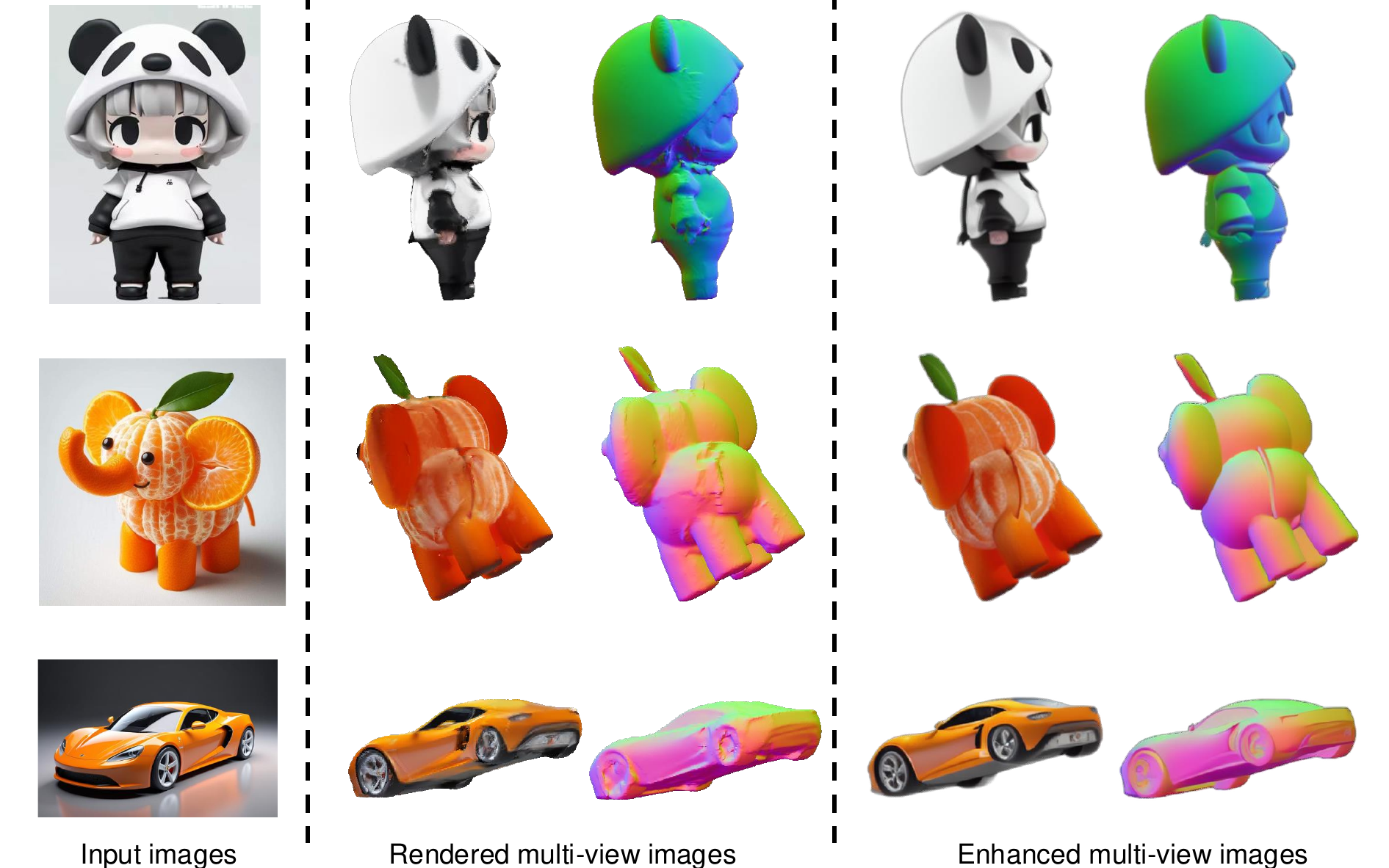}
\caption{Visualization of our cross-domain multi-view enhancement method.}
\vspace{-4mm}
\label{fig:supp_enhance}	
\end{figure}

%% file: figures/supp_mesh_compare.tex
\begin{figure*}[t]
\centering
\includegraphics[width=0.9\linewidth]{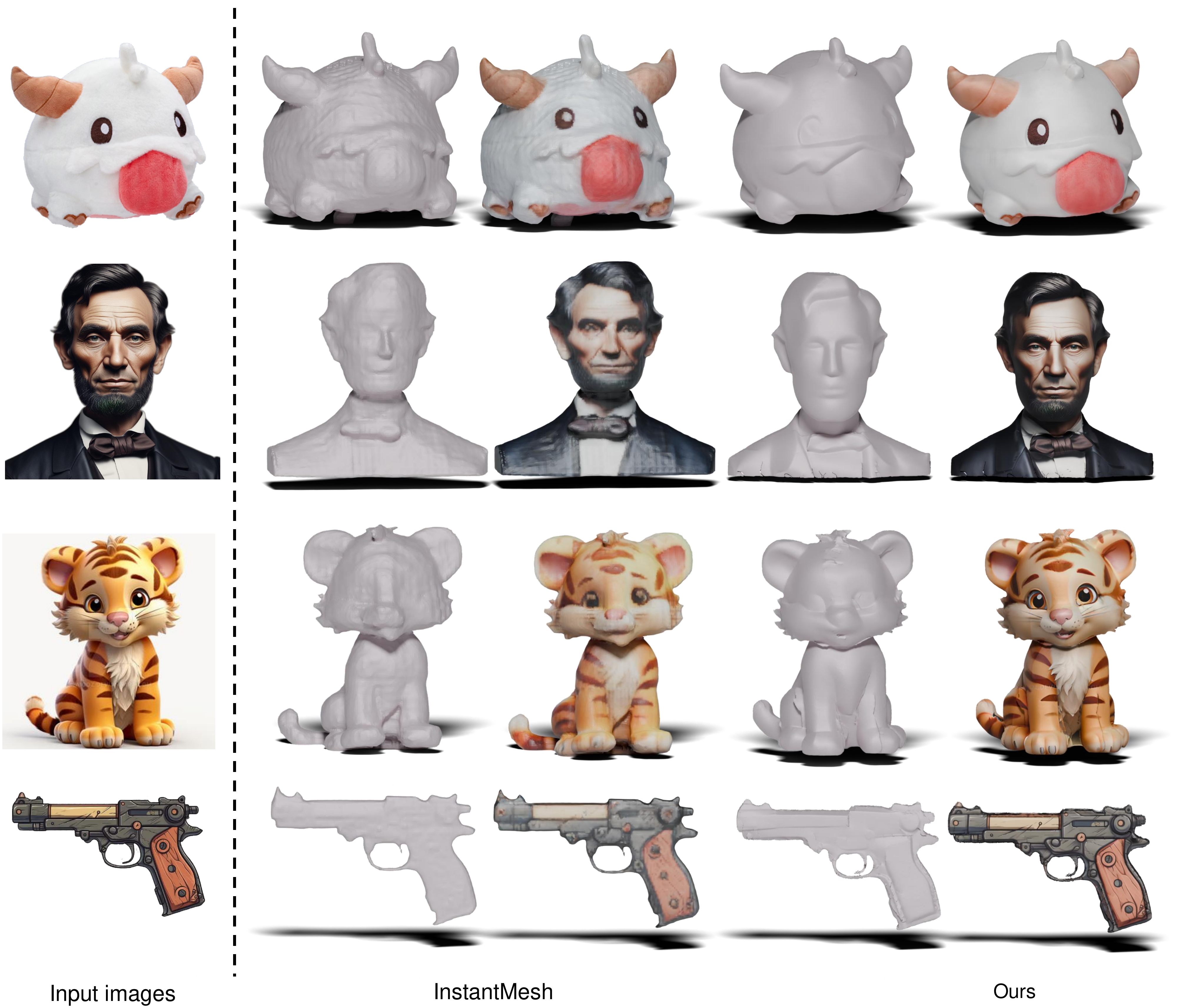}
\caption{Qualitative comparison on single image to 3D task between InstantMesh~\cite{xu2024instantmesh} The mesh generated by our method not only has better geometric and texture details but also has higher consistency with the input image.}
\label{fig:supp_mesh_compare}	
\vspace{-2mm}
\end{figure*}

%% file: figures/supp_nvs.tex
\begin{figure*}[t]
\centering
\includegraphics[width= \linewidth]{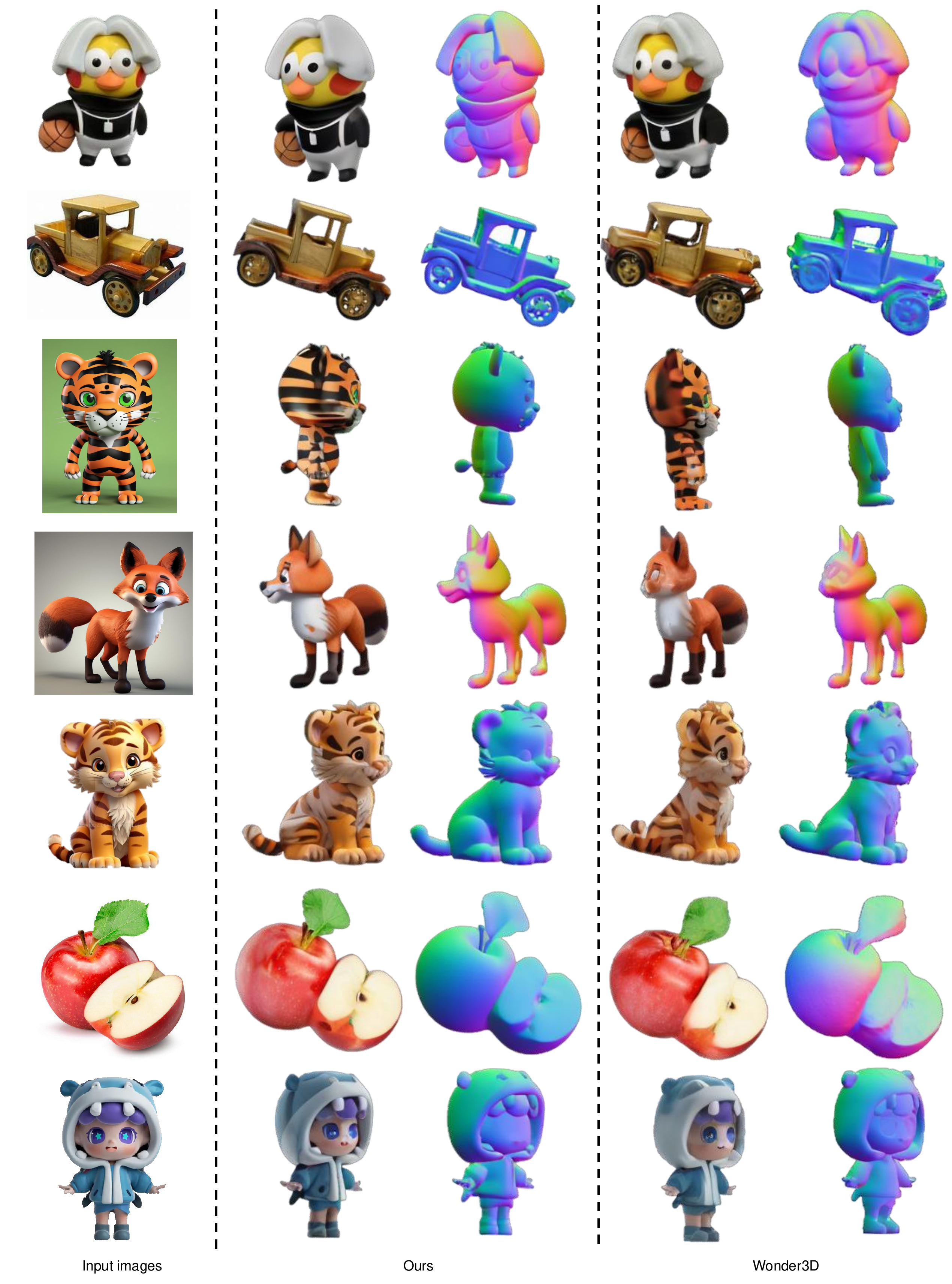}
\caption{Qualitative comparison on cross-domain NVS task with Wonder3D~\cite{long2024wonder3d}. Our method generates more fidelity rgb images and sharper normal maps. }
\label{fig:supp_nvs}	
\end{figure*}